\documentclass{ieeeaccess}
\usepackage{amsmath,amsfonts,amssymb}
\usepackage{algorithm}
\usepackage{algpseudocode}
\usepackage{array}
\usepackage[caption=false,font=normalsize,labelfont=sf,textfont=sf]{subfig}
\usepackage{textcomp}
\usepackage{stfloats}
\usepackage{url}
\usepackage{verbatim}
\usepackage[dvipdfmx]{graphicx}
\usepackage{cite}
\usepackage{bm}
\usepackage{nomencl}
\usepackage{multirow}
\usepackage{color}

\newcommand{\ocut}[1]{}

\renewcommand{\algorithmicrequire}{\textbf{Input:}}
\renewcommand{\algorithmicensure}{\textbf{Output:}}

\graphicspath{{./imgs/}}

\newcommand{\jtextd}[1]{}  

\makenomenclature

\newcounter{const} 
\newcounter{repair}

\makeatletter
\AtBeginDocument{\DeclareMathVersion{bold}
\SetSymbolFont{operators}{bold}{T1}{times}{b}{n}
\SetSymbolFont{NewLetters}{bold}{T1}{times}{b}{it}
\SetMathAlphabet{\mathrm}{bold}{T1}{times}{b}{n}
\SetMathAlphabet{\mathit}{bold}{T1}{times}{b}{it}
\SetMathAlphabet{\mathbf}{bold}{T1}{times}{b}{n}
\SetMathAlphabet{\mathtt}{bold}{OT1}{pcr}{b}{n}
\SetSymbolFont{symbols}{bold}{OMS}{cmsy}{b}{n}
\renewcommand\boldmath{\@nomath\boldmath\mathversion{bold}}}
\makeatother

\def\BibTeX{{\rm B\kern-.05em{\sc i\kern-.025em b}\kern-.08em
    T\kern-.1667em\lower.7ex\hbox{E}\kern-.125emX}}

\begin{document}
\history{Date of publication xxxx 00, 0000, date of current version xxxx 00, 0000.}
\doi{10.1109/ACCESS.2024.0429000}

\title{Vehicle Painting Robot Path Planning Using Hierarchical Optimization}
\author{\uppercase{YUYA NAGAI}\authorrefmark{1}, 
\uppercase{HIROMITSU NAKAMURA}\authorrefmark{2}, 
\uppercase{NARITO SHINMACHI}\authorrefmark{2},
\uppercase{YUTA HIGASHIZONO}\authorrefmark{2},
\uppercase{SATOSHI ONO}\authorrefmark{1},
}

\address[1]{Department of Information Science and Biomedical Engineering,
Graduate School of Science and Engineering, Kagoshima University\\
1-21-40, Korimoto, Kagoshima, 890-0065 Japan \\}
\address[2]{TOYOTA Body Research \& Development Co., Ltd,
395-1, Uenodan, Kokubu, Kirishima, 899-4461 Japan\\}

\markboth
{Nagai \headeretal: Vehicle Painting Robot Path Planning Using Hierarchical Optimization}
{Nagai \headeretal: Vehicle Painting Robot Path Planning Using Hierarchical Optimization}

\corresp{Corresponding author: SATOSHI ONO (e-mail: ono@ibe.kagoshima-u.ac.jp).}

\begin{abstract}
In vehicle production factories, the vehicle painting process
employs multiple robotic arms to simultaneously apply paint to car
bodies advancing along a conveyor line.
Designing paint paths for these robotic arms, which involves assigning
car body areas to arms and determining paint sequences for each arm,
remains a time-consuming manual task for engineers, indicating the
demand for automation and design time reduction.
The unique constraints of the painting process hinder the direct
application of conventional robotic path planning techniques, such as
those used in welding.
Therefore, this paper formulates the design of paint paths as a
hierarchical optimization problem, where the upper-layer subproblem
resembles a vehicle routing problem (VRP), and the lower-layer
subproblem involves detailed path planning.
This approach allows the use of different optimization algorithms at
each layer, and permits flexible handling of constraints specific to
the vehicle painting process through the design of variable
representation, constraints, repair operators, and an initialization
process at the upper and lower layers.
Experiments with three commercially available vehicle models
demonstrated that the proposed method can automatically design paths
that satisfy all constraints for vehicle painting with quality
comparable to those created manually by engineers.
\end{abstract}


\titlepgskip=-21pt

\maketitle

\section{Introduction}
\begin{figure}[t]
  \centering
  \includegraphics[width=0.45\textwidth]{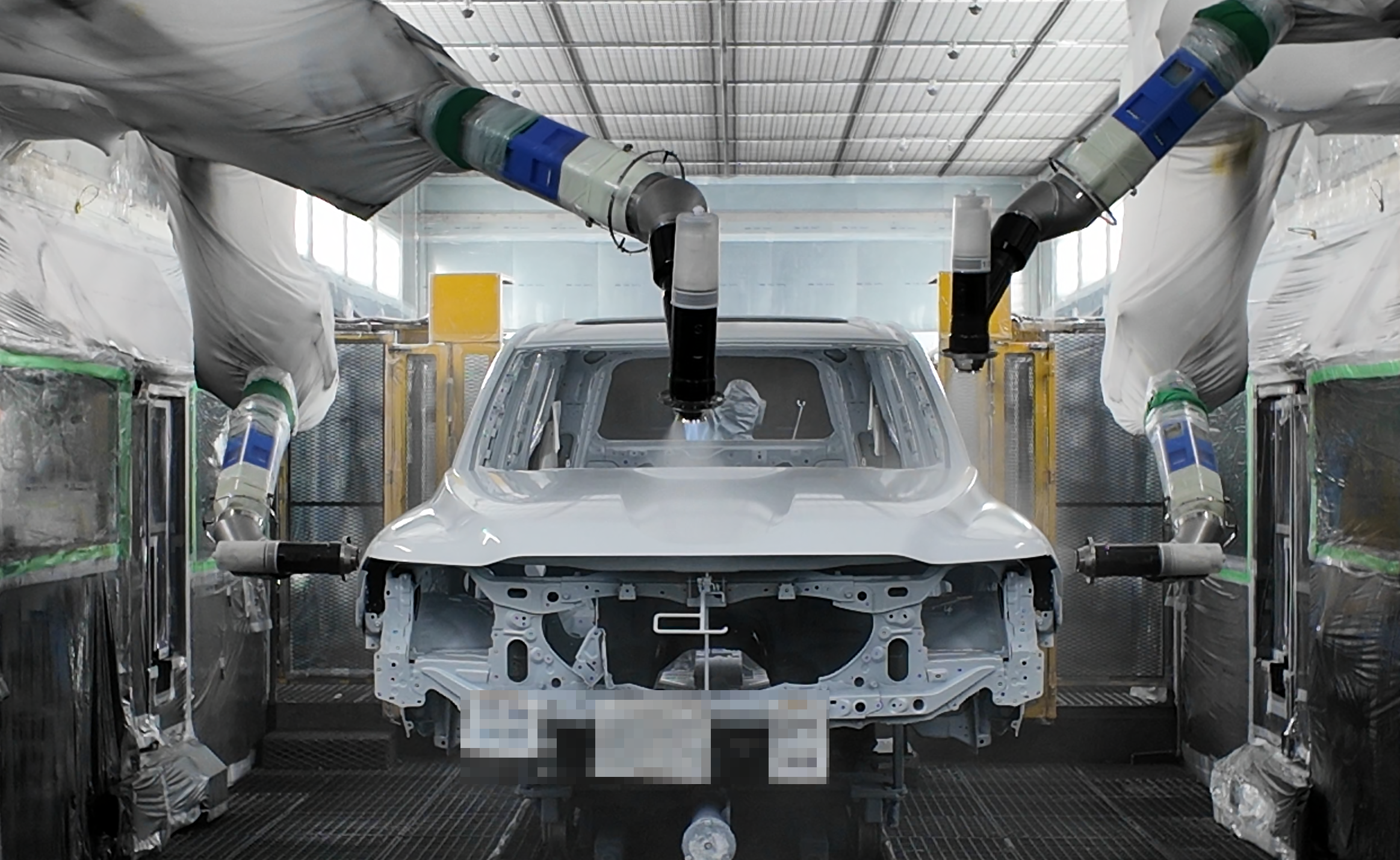}
  \caption{Vehicle painting by multiple robot arms\protect\footnotemark[3] .}
  \label{fig:factory}
\end{figure}

Automobile production involves various stages, including steel plate
pressing, body welding and painting, and assembly of powertrains and
interiors.
During the painting process, multiple robotic arms spray paint onto
vehicle bodies traveling along a production line.
As shown in Fig.~\ref{fig:factory}, three to four robots are positioned on
each side of the line to paint a hood, fenders, doors, and a rear door.
Depending on the factory or vehicle model, a roof may also be painted
simultaneously.
For each panel (e.g., a door or a fender) that forms the vehicle body,
these robot arms move in a repetitive back-and-forth motion while
spraying paint from the gun mounted on their arm heads, thereby
achieving a uniform painting.

\footnotetext[3]{\url{https://global.toyota/jp/company/plant-tours/painting/#sec09}}

Although the spraying process itself is automated, the design of these
paint paths is carried out by skilled engineers relying on experience
and intuition.
Designing the painting paths for robotic arms comprises three major
steps.
First, fragments of paint paths required to comprehensively cover the
entire vehicle are created, based on the vehicle's shape and the
spray range of the nozzle attached to each robotic arm head.
Next, these paint path fragments are allocated to robot arms, and the
painting sequence is determined.
Finally, detailed path planning and validation, including arm posture
determination, are conducted in a simulator according to the chosen
painting sequence.

\begin{figure}[t]
  \centering
  \includegraphics[width=0.35\textwidth]{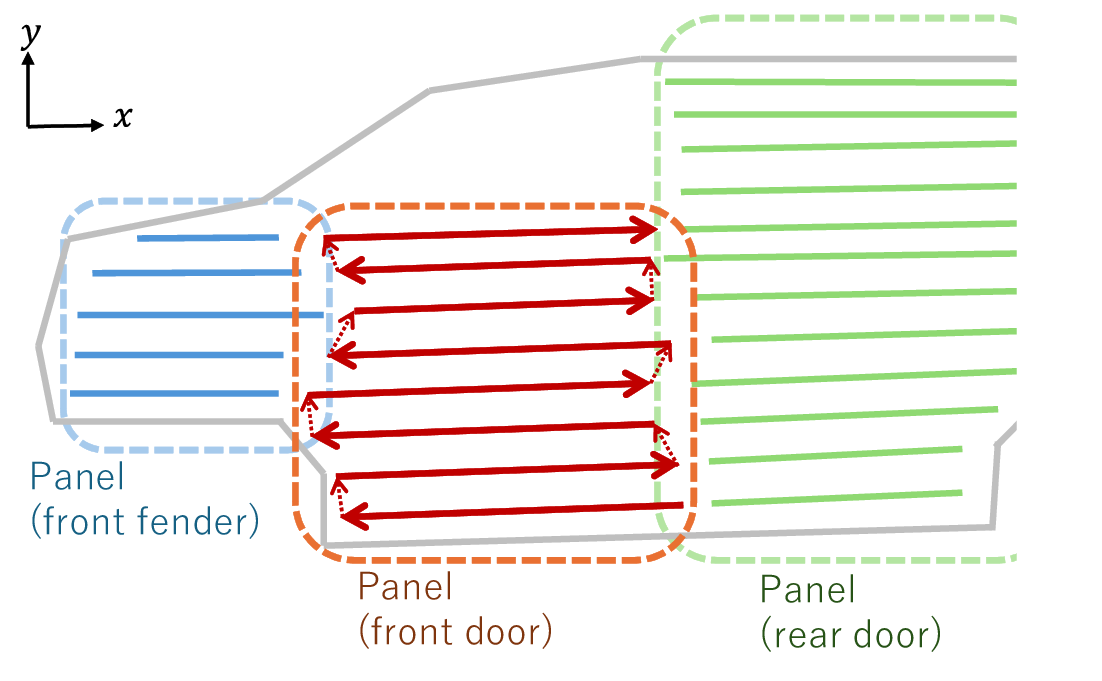}
  \caption{Example process of painting a panel (front door) from
    bottom to top with lateral back-and-forth movement.}
  \label{fig:ex_paint_panel}
\end{figure}

These processes require addressing paint-specific constraints,
including completing the task within the area reachable by the moving
vehicle and controlling the painting sequence for quality, as well as
standard constraints for robotic arms, such as remaining within their
range of motion and avoiding inter-arm collisions.
Constraints for maintaining paint quality require that the
arm moves at a constant speed and that, for
vertically installed panels (e.g., fenders and doors), the painting
process proceeds upward in a continuous manner, achieved by lateral
back-and-forth movement, as shown in Fig.~\ref{fig:ex_paint_panel}.
Further constraints exist to simplify equipment and process
management, such as standardizing painting heights across arms and
ensuring that paint emitted from one gun does not adhere to other guns
to keep the spray guns as clean as possible.

\begin{figure*}[t]
  \centering
  \includegraphics[width=0.95\textwidth]{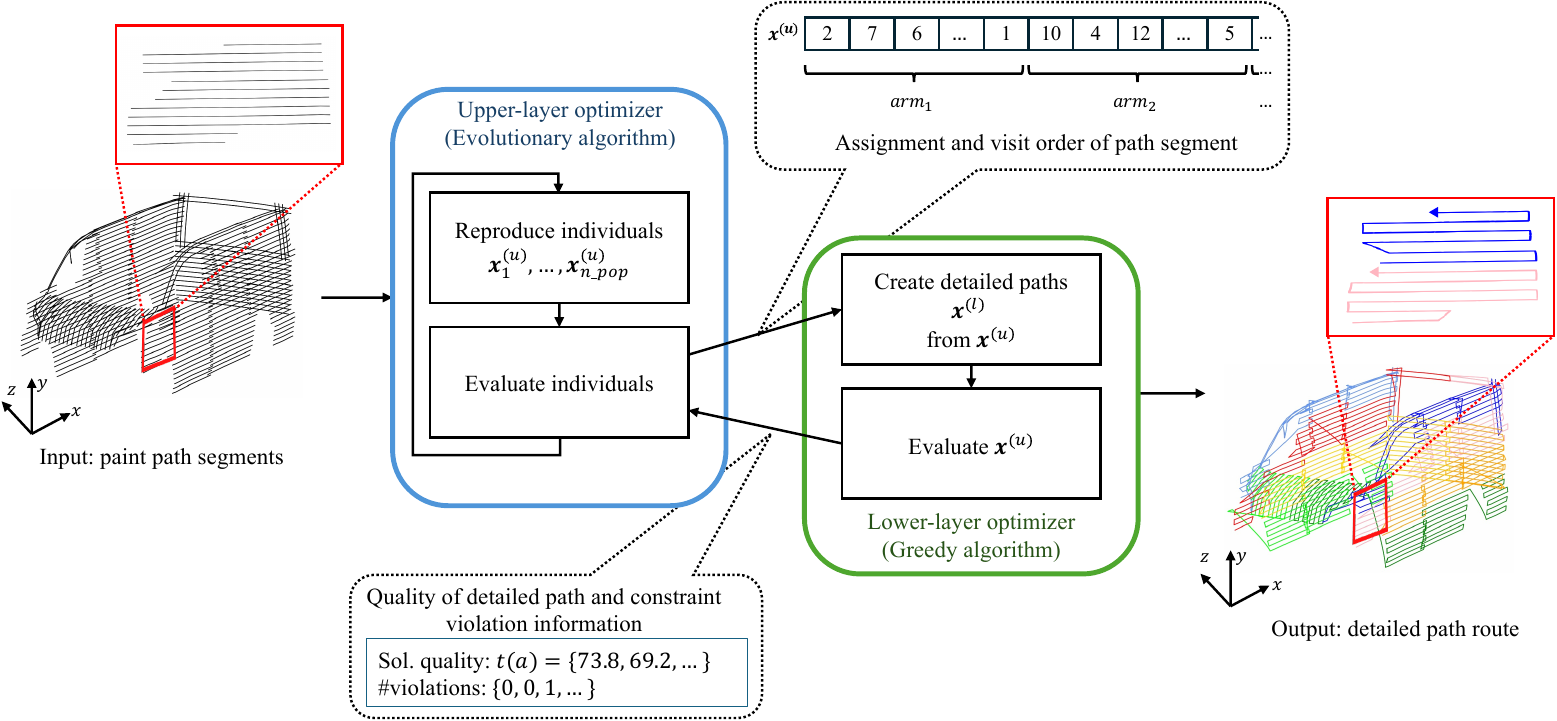}
  \caption{Overview of the proposed method.}
  \label{fig:overview}
\end{figure*}

Currently, these tasks for paint path design are performed based on
the expertise of skilled engineers, ensuring that vehicles moving
along the line remain within each arm's operational range and that
the workload among arms is balanced.
Through trial and error, they assign paint areas and
determine painting sequences.
When there are significant differences in vehicle body shape or factory
environments, such as the number of robotic arms or their arrangements, 
an entirely new path planning is required.
Even when the body shape resembles an existing model, adjustments to
the existing plan are often insufficient to achieve high-quality
paths, necessitating further trial and error.

Numerous studies have extensively investigated path planning for
multiple robotic arms on general tasks that do not involve vehicle
body painting.
For instance, studies have focused on tasks such as object
grasping~\cite{tika2020optimal,shome2021roadmaps} and welding on
automobile bodies~\cite{spensieri2015iterative, baizid2015time,
  touzani2021multi}.
However, many of these studies examine tasks in which robotic arms
move to a designated position and perform operations while
stationary.
Although research has begun to address tasks where robotic arms
operate while in motion, such as
painting~\cite{zbiss2022automatic,ortiz2024routing}, it remains
underdeveloped compared to welding path planning.
In particular, constraints necessary for maintaining paint quality are
crucial for practical applications; however, as far as we know, these
existing research does not specifically consider these constraints.

This paper proposes a method that models the design of paint paths for
vehicle bodies using multiple robotic arms as a hierarchical
optimization problem.
The approach divides the task into an upper-layer subproblem, which
allocates paint regions to each robotic arm and establishes their
painting schedule, and a lower-layer subproblem which plans the detailed
paths for each arm, as shown Fig.~\ref{fig:overview}.
The proposed method formulates the upper-layer subproblem as a variant of
a Vehicle Routing Problem (VRP) and solves it using an evolutionary
algorithm, while addressing the lower-layer subproblem by constructing a
simplified greedy optimizer tailored to vehicle painting.
Unlike welding, where the arm head is fixed at the work point, 
painting requires the arm head to move while spraying paint.
This difference complicates the formulation of the assignments of
paint areas to the arms and the determination of the painting order as
a canonical VRP.

\begin{figure}[t]
  \centering
  \begin{tabular}{@{}p{0.235\textwidth}@{~~}p{0.235\textwidth}@{}}
    \includegraphics[width=0.235\textwidth]{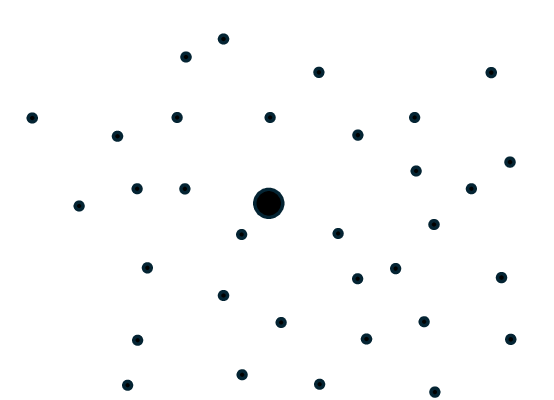} &
    \includegraphics[width=0.235\textwidth]{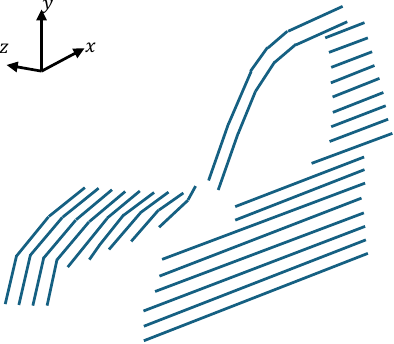} \\
    {\footnotesize (a) Customers to be visited in VRP} &
    {\footnotesize (b) Paint path segments to be visited in vehicle body painting}
  \end{tabular}
  \caption{Difference of customers (visiting points) between canonical VRP and vehicle body painting.}
  \label{fig:city_shape}
\end{figure}

The proposed method treats a line segment of a painting path as a customer
in the VRP, as shown in Fig.~\ref{fig:city_shape}, because the paint gun
attached to the arm head ideally moves linearly from one end of a
panel to the other.
Whereas canonical VRP treats customers as dimensionless points, 
the painting problem represents them as line segments.
Consequently, even after determining the visiting order of segments, the
starting direction for painting each segment remains ambiguous,
making
it difficult to define a precise path of each arm head.
Thus, the proposed method
requires the lower-layer optimizer to
specify the detailed movements of the paint arm head based on VRP-based
solutions.
This process not only determines the starting direction for each path
segment corresponding to each customer in the VRP but also designs the
transition paths between segments and any necessary waiting times to
satisfy constraints.

The adoption of hierarchical problem decomposition and evolutionary
algorithms allows the proposed method to address various constraints
specific to the vehicle painting task through mechanisms such as
variable representation, penalty functions, and repair operators at
the upper layer, as well as rules of the greedy optimizer at the lower
layer.
For instance, a constraint requires continuous painting from bottom to
top within a single panel, such as a door or fender.
The proposed method addresses this constraint by introducing a repair
operator in the upper-layer optimizer that modifies the solution.
Additionally, a constraint that requires each arm to complete painting
its assigned area within its operational range while the vehicle is
moving is verified in the lower-layer optimizer.
If this constraint is not satisfied, a penalty is added to the
objective function value of the upper-layer solution.

Experiments were performed on three vehicle models produced at two
factories, demonstrating that the proposed method can account for
diverse constraints in designing paths for multiple robotic arms and
produce paths comparable to those designed by skilled engineers.

The main contributions of this paper are as follows:
\begin{itemize} 

\item {\bf The first practical method for vehicle painting robot path planning 
that consider constraint of painting quality and equipment maintenance:}
To the best of our knowledge, no previous work has tackled the problem
of designing paths for painting a moving vehicle body using multiple
robot arms while satisfying production-grade paint quality
requirements in a real manufacturing environment.
We clarify these constraints and assign them a two-tier priority,
thereby enabling feasible solutions that consider the arms' physical
limitations, paint quality, and the impact on production equipment.

\item {\bf Formulation that facilitates deriving solutions through the
  combination of established algorithms:}
Our method adopts standard techniques from previous multi-robot path
planning research, treating the vehicle painting problem as a
hierarchical optimization challenge. 
In particular, by formulating the upper-layer subproblem in a VRP-like
style, we reduce the specialized challenges of painting tasks and
show that standard evolutionary computation approaches are applicable.
We also demonstrate that a simple GA combined with a greedy algorithm
generates solutions of sufficiently high quality.
This structure holds importance because it supports the future
application of various techniques and algorithms accumulated in the
field of VRP~\cite{kumar2012survey,braekers2016vehicle,zhang2022review}.

\item {\bf Introducing various constraint handling mechanisms:}
Our method achieves effective constraint handling through problem
decomposition and the use of evolutionary algorithms. 
Specifically, we maintain the conventional genetic algorithm framework
while incorporating the various paint-specific constraints into
the optimization process via combined mechanisms involving
solution representation, penalty functions, repair operators, improved
initial solution generation.

\end{itemize}

This paper is organized as follows:
Section \ref{sec:related_work} reviews studies on path planning for
multiple robotic arms in automotive body welding and painting
processes.
Section \ref{sec:proposed_method} describes the proposed method for
paint path design, including its structure and process flows.
Section \ref{sec:experiments} evaluates the effectiveness of the
proposed method through experiments using actual vehicle and factory
data.

\section{Related work}
\label{sec:related_work}

Research on path planning for robotic arms has been conducted
extensively, depending on the number and types of robotic arms and the tasks they are intended to
perform~\cite{xidias2010time,wang2019new,tika2020optimal,shome2021roadmaps,touzani2021multi,mcgovern2023general}.
This section presents a review of related studies, focusing on welding
and painting as key tasks that require path planning for multiple
robotic arms in automotive body production.

\subsection{Studies focused on welding processes}

Spensieri et al. proposed a method for welding path design that
divides the problem into two steps: collision-free and collision-aware
path designs~\cite{spensieri2015iterative}.
They formulated the collision-free path design as a min-max multiple
generalized traveling salesman problem (MGTSP) to minimize the
operating time of the arm with the heaviest workload by determining
the sequence in which each arm (salesman) visits welding points
(cities).
They also formulated the collision-aware path design
as a generalized
traveling salesman problem (GTSP) with synchronization constraints
among all arms, ensuring that all arms operate on common time steps, thus
avoiding collisions.

Touzani et al. proposed a method to design robot paths using
synchronization signals to avoid collisions, achieving paths with
reduced cycle time in simulations and actual robots\cite{9681247}.
This method also employs an approach that divides the problem into
subproblems, first optimizing each robot's path, and then calculating
synchronization signals to prevent robots from simultaneously occupying
collision-prone areas. 
Unlike Spensieri's work~\cite{spensieri2015iterative}, this method
formulates the problem as a min (sum-max) MGTSP to minimize both the
total cost of all arms and the highest individual arm cost, thereby
considering not only cycle time but also the workload of each arm.
The methods solveseach robot's path design using a Genetic Algorithm
(GA) that minimizes the total distance between visited points.
When the generated path results in an obstacle collision, this method
uses a probabilistic roadmap (PRM) to generate alternative waypoints,
ensuring a collision-free trajectory.
If a collision between arms arises, this method mitigates the risk of
multiple arms entering collision-prone zones by designating one arm to
wait.
To facilitate this, they formulate a problem that minimizes waiting
time and then resolve it using a graph-based
method~\cite{spensieri2013coordination, kavraki2016motion}.

Pellegrinelli et al. proposed a spot-welding cell design method that determines
both the types and arrangements of arms alongside path
design~\cite{pellegrinelli2017multi}.
This method divides the unified problem into four stages.
Stage 1 involves designing a single arm's path using PRM to avoid
collisions with obstacles.
Stage 2 assesses the potential for collisions between arms based on
these single-arm paths.
For collision detection, this method utilizes the volume-swept-like
algorithm ~\cite{peternell2005swept}, which generates point clouds
from arm shapes and movement trajectories.
Stage 3 involves optimizing arm placement and designing paths that
prevent collisions between arms.
For each configuration, robot paths are designed by formulating the problem as
a traveling salesman problem (TSP) and minimizing the cost calculated
based on the placement of the arms.
Stage 4 involves verifying the paths created in Stage 3 by simulation to
ensure that no collisions occur with obstacles or between arms.
They applied their method to a welding task involving three arms and
succeeded in determining robot arm placements and generating their
paths that reduced the cycle time below the specified threshold.

\subsection{Studies focused on painting processes}

Asakawa et al. proposed a path planning method for a single arm that
does not require specialized knowledge of the painting
process~\cite{asakawa1997teachingless}.
This method designs a path using a computer-aided design (CAD) model
of a three-dimensional (3D) car body shape, and generates control
commands to operate the arm based on the path information, and
employs over-splay points to address the problem at the edges of the
workpiece.

Path planning using multiple arms requires accounting for inter-arm
collisions, making it more complex than single-arm path planning.
Additionally, painting tasks involve moving the arms during operation,
which increases the difficulty compared to designing paths for welding
arms.
Zbiss et al. introduced a path planning approach that employs multiple
robotic arms to paint a stationary chassis~\cite{zbiss2022automatic}.
This method decomposes the painting path planning problem into
multiple sub-problems, thereby enabling the design of constrained
paths.
The method employs
the Boustrophedon decomposition
algorithm~\cite{choset1998coverage,choset2000coverage} to generate
two-dimensional (2D) paths for each component and then extends them to
three-dimensions (3D)
by connecting the paths between components.
During the design process, the method identified collision-prone and
collision-free regions based on the operating range to prevent
collisions between robotic arms.

In actual manufacturing settings, vehicle body painting takes place
either when the vehicle is stationary on the production line or while
it moves along the line.
Movement along the line causes changes in the paintable areas
within each arm's range of motion over time, thereby limiting the time
available for painting specific sections. 
Shahputra et al. investigated the process in which vehicles move along
the production line~\cite{syahputra2018car}.
They proposed a method that models the assignment of robots to paint
vehicle models in a specific order during the painting process,
employing a flow scheduling shop framework and solving it with a
harmony search algorithm.
This approach yields task sequences that reduce the overall painting
cycle time.

Paint quality is an essential element in the automotive
body painting process~\cite{sheng2001cad,chen2008automated}.
Achieving a uniform paint thickness is one of the critical quality
criteria.
Trigatti et al. introduced a path planning method to improve paint
uniformity for the surfaces of general industrial
products~\cite{trigatti2018new}.
Following the approach of previous
research~\cite{ramabhadran1997fast,andulkar2015automated}, they first
established an end-effector path based on an object's shape and
designated painting areas, and then developed a complete path,
including the arm joints, using inverse kinematics.
Additionally, by imposing limits on the end-effector's speed and
acceleration and smoothing the arm's motions, they reduced paint
inconsistencies and achieved a uniform paint thickness.

In addition to ensuring uniform paint thickness, maintaining paint
quality requires that the arms to continuously paint adjacent regions.
In scenarios with a single arm or when painting a stationary vehicle,
continuous areas of each component are painted from the lower or upper
sections to optimize cycle time, as demonstrated in
studies~\cite{tang2015surface,zbiss2022automatic,gleeson2022generating}.
However, when utilizing multiple arms to paint a vehicle moving along
the production line, painting each component from the lower or upper
sections can cause the painted areas to move outside the arms'
operational range during the process. 
Designing routes for multiple robot arms to paint a moving vehicle
body, while ensuring the aforementioned high paint quality, poses a
significant challenge.

\section{The proposed method}
\label{sec:proposed_method}

\subsection{Key idea}
\label{ssec:key_idea}

This study focuses on paint path design for a vehicle body moving on the
production line.
The problem involves assigning path segments to each robotic arm,
determining the paint order, and designing detailed paths based on the
determined sequence.
Although it is necessary to consider the orientation of the arm heads,
we assume that the arm orientation is basically determined by the
panel and therefore do not include it as an optimization target.
Instead, we only account for the cost associated with significant
changes in arm orientation when moving between panels.

The proposed method formulates this problem as a hierarchical problem
involving the upper-layer combinatorial optimization problem similar
to a VRP and the lower-layer subproblem for detailed path planning.
This approach method has two main advantages:
\begin{enumerate}
  \item The hierarchical approach facilitates the use of
    meta-heuristics to discover near-optimal solutions within
    practical time by leveraging evolutionary algorithms developed for
    VRP and the insights from previous studies in the upper-layer
    solver.
    This study employs a simple greedy optimizer to solve the
    lower-layer subproblem.
    When employing the proposed method for practical implementation, a
    higher-quality simulator currently used by a manufacturer can be
    used without the need to redesign the upper solver.

  \item Combining hierarchical formulation and evolutionary algorithms
    enables the proposed method to address constraints in various ways
    during formulation and optimization, thereby considering more
    constraints than conventional studies, as shown in
    Table~\ref{table:constraints_compare}.

\end{enumerate}

\begin{table*}[t]
  \caption{Differences in conditions considered by the preivous methods and the proposed method\protect\footnotemark[4] .}
        \label{table:constraints_compare}
        \centering
        \begin{tabular}{l|l|ccccc}
        \hline 
        & & \multicolumn{4}{c}{Constraints to consider} \\ \cline{3-7}
        Method & Task & Motion range  & Multiple &Collisions   & Movement of  & Paint   \\
               &      & of robot arms & arms     &between arms & vehicle body & quality \\
        \hline
        Spensieri et al.~\cite{spensieri2015iterative} & Welding 
        & \checkmark & \checkmark & \checkmark  &  &   \\
        Zbiss et al.~\cite{zbiss2022automatic} & Painting    
        & \checkmark & \checkmark & \checkmark  &   &\\
        Shahputra et al.~\cite{syahputra2018car} & Painting    
        & \checkmark & \checkmark & \checkmark  & \checkmark  &\\ 
        Tang et al.~\cite{tang2015surface} & Painting  
        & \checkmark & \checkmark & \checkmark  &   &\\
        Gleeson et al.~\cite{gleeson2022generating} & Painting
        & \checkmark &     &   &  & \checkmark\\
        The proposed method & Painting
        & \checkmark & \checkmark & \checkmark & \checkmark & \checkmark\\
        \hline
        \end{tabular}
        \\
\end{table*}

\footnotetext[4]{%
Paper~\cite{ortiz2024routing} is not included in Table~\ref{table:constraints_compare}
because it was difficult to obtain details about the methods and
experimental results from the paper.
}

\subsection{Overview of the proposed method}
\label{ssec:overview}

Fig.~%
\ref{fig:overview}
shows the structure and the process flow of
the proposed method.
Similar to the previous method~\cite{zbiss2022automatic}, the proposed
method divides the vehicle body painting path planning problem into two
subproblems.
Specifically, it models the problem by combining a combinatorial
optimization problem resembling a Vehicle Routing Problem
(VRP)~\cite{dantzig1959truck}, 
which determines the visiting sequence of painting locations
(analogous to customers) and assigns robotic arms, with a subproblem
that derives detailed paths from the established visit order and arm
assignments.

This paper defines a painting path segment as a short linear path
where the painting is conducted with fixed gun directions, angles,
speeds, and spray volumes.
The input of the problem is a set of painting path segments, which can be
easily defined according to a target 3D shape of a vehicle body, as
shown in Fig.~\ref{fig:overview}.
The proposed method consists of a meta-heuristic-based optimizer that
addresses the upper-layer subproblem and a greedy optimizer that handles
the lower-layer subproblem using the solver-generated candidate
solutions.
The solver employs a Genetic Algorithm (GA) for optimization, while
the simulator utilizes a greedy algorithm. 
GA basically iterates candidate solution generation and evaluation.
During the evaluation,
the simulator generates detailed paths by solving the subproblem and
calculates the objective function values of the candidates based on
these designed paths.

\begin{figure}[t]
  \centering
  \includegraphics[width=0.48\textwidth]{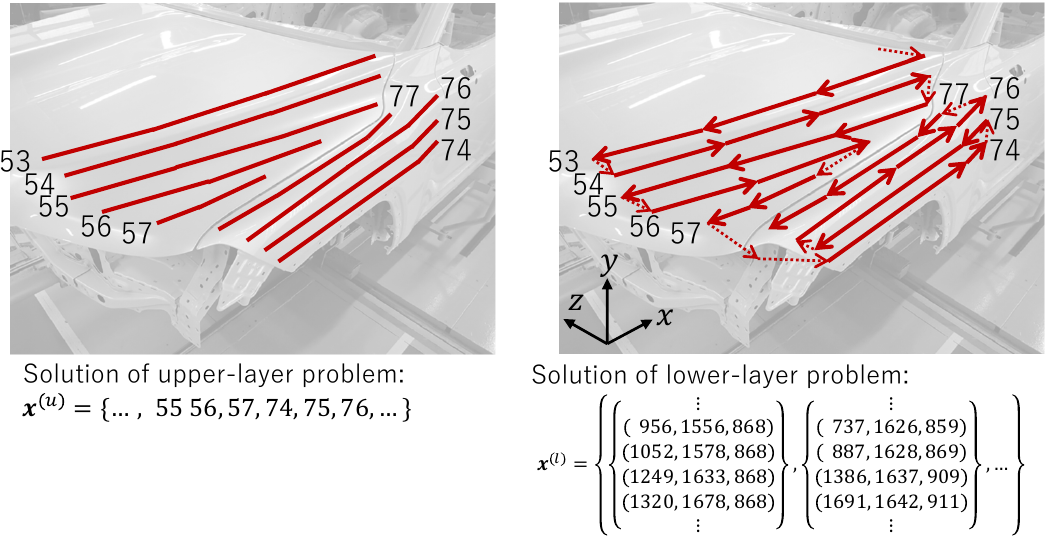}
  \caption{Example solutions of upper- and lower-layer subproblems.}
  \label{fig:solutions}
\end{figure}

The proposed method addresses a variety of constraints, including
those specific to the painting process, by individually handling each
constraint's characteristics within the solver and simulator
processes.
For instance, painting requires robotic arms to spray paint while
moving linearly.
Unlike path planning for welding robotic arms in related studies,
where welding points can be directly represented as customers in a
problem such as VRP and TSP,
applying a similar approach to this problem is challenging.
Therefore, our method models customers as straight or curved painting
path segments. 
This enables uniform painting by moving linearly and avoiding
paint unevenness.
The upper-layer subproblem determines only the sequence of painting path
segment visits, while the lower-layer subproblem determines from which
endpoint to begin painting each segment, as shown in Fig.~\ref{fig:solutions}.

\subsection{Formulation}

\begin{figure}[t]
  \centering
  \includegraphics[width=0.48\textwidth]{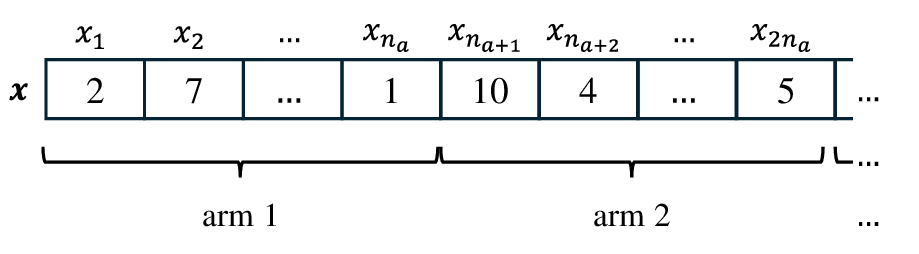}
  \caption{Genotypic representation in the upper-layer subproblem.}
  \label{fig:general_ind}
\end{figure}

\subsubsection{Variables}
\label{sssec:variables}

The goal of this task is to determine the detailed trajectories for
$n_{arms}$ robot arm heads, $\{ \bm{r}_{a, t} \}_{a \in \{1, \ldots,
  n_{arms}\}, t \in \{1, \ldots, t_{max}\}}$, so that they can paint
one vehicle body in the shortest possible time.
As described in Section~\ref{ssec:key_idea}, the proposed method
decomposes the problem into an upper-layer and a lower-layer
subproblems. 
Specifically, the upper-layer subproblem involves determining
which robot arm will paint each segment of the painting path and in
what order, with the solution denoted by $\bm{x}^{(u)}$. 
The lower-layer solver then focuses on finding the detailed
trajectories, with its solution represented as $\bm{x}^{(l)}$,
that is, 
\begin{equation}
\bm{x}^{(l)} = \{ \bm{r}_{a, t}  \}_{a \in \{1, \ldots, n_{arms}\}, t \in \{1, \ldots, t_{max}\}}
\end{equation}
If we consider the overall solution, which incorporates both the
upper- and lower-layer results, we represent it by $\bm{x}$.

To facilitate the application of existing metaheuristics such as GA, we
formulate the upper-layer subproblem as a variant of VRP and adopt a
solution representation similar to that of VRP~\cite{prins2004simple,
  baker2003genetic, nazif2012optimised}.
Specifically, a candidate solution, represented by an individual
$\bm{x}^{(u)}$ is a permutation of paint path segment identification numbers
(IDs) arranged in the order of visits as shown in
Fig.~\ref{fig:general_ind}.
The total number of design variables $n_{dim}$ equals the total number of segments $n_{segs}$,
and each variable element $x_i$ assigns a segment ID using integers
from 1 to $n_{dim}$ ($x_i \in \{1, \ldots, n_{dim} \}$).
Both VRP and the current problem involve determining the routes of
multiple trucks or robotic arms.

For simplicity, we assume that each robotic arm visits an equal number
of customers and represent both the assignment and the visit order of
painting locations (customers) as described below.
\begin{equation}
\bm{x}^{(u)} = \left\{ x_1, x_2, \ldots, x_{n_{dim}} \right\}, ~~ (x_i \neq x_j,~ \forall~ i,j \in \{1, \ldots, n_{dim}\})
\end{equation}
That is, $x_1$ to $x_{y/n_{arms}}$ indicate the path segments
(corresponding to customers in VRP) and their visitation order for the first robot arm, while
$x_{n_{dim}/n_{arms}+1}$ through $x_{2n_{dim}/n_{arms}}$ denote those for the second robot arm.
Thus, the assignment of customers to each robot arm relies on the positions of elements in $\bm{x}^{(u)}$
Fig.~\ref{fig:general_ind} shows an example of a solution candidate of
the upper-layer subproblem, which corresponds to the genotypic
representation of GA.
Each robot arm takes on $n_a$ path segments, and $arm_1$ visits the
path segments sequentially: 2, 7, 6 and others.

Fig.~\ref{fig:solutions} illustrates example solutions for both the
upper-layer and lower-layer subproblems.
The figure, which focuses on the boundary between an vehicle body's
hood and front fender, shows that the upper-layer solution is a
permutation of the given painting path segment IDs, while the
lower-layer solution consists of the detailed trajectory of a robot
arm's head.
Although the figure displays only a portion of one robot arm's
trajectory, the full problem involves designing trajectories for
multiple robot arms simultaneously.

In the previous discussion, we assumed that each arm would receive an
equal number of painting path segments; 
however, in practice, allowing variability in the segment allocation
can reduce the overall work time.
Therefore, the proposed method introduces dummy segments $s_{1}^{(d)},
\ldots, s_{n_d}^{(d)}$ that, together with the standard painting
segments, are assigned to each arm.
This approach allows for an uneven distribution while preserving a
simple formulation, and the total number of variables $n_{dim}$ for the
upper-layer subproblem equals $n_{segs} + n_d$.

\subsubsection{Objective function}

The objective of this problem is to minimize the work time required to
paint an vehicle body while satisfying all constraints, i.e., 
\begin{eqnarray}
   \mbox{minimize} && \quad  f(\bm{x}^{(l)}) = \max_{a} t_a(\bm{x}^{(l)})\\
   \mbox{subject~to} 
   && \quad  g_c(\bm{x}^{(l)}) \geq 0 
   \qquad \forall c \in \mathbb{C}
\end{eqnarray}
where $t_a(\cdot)$ denotes the work time of arm $a$, 
$g_i(\cdot)$ denotes constraint functions, and
$\mathbb{C}$ denotes a set of constraints.
The work time $t_a(\cdot)$ for arm $a$ is calculated from the moment the
vehicle body's front reaches a predetermined reference position until
the arm completes painting all its assigned segments and returns to
its starting point.

The upper-layer optimizer in our proposed method finds a solution that 
minimizes the objective function $f(\cdot)$ while satisfying all constraints:
\begin{equation}
   \mbox{minimize}  \quad  f(\bm{x}^{(u)}) = \max_{a} t_a(\bm{x}^{(l)})
   + \sum_q p_q (\bm{x}) 
\end{equation}
where
$p_q(\cdot)$
denotes penalty functions.
The objective function is defined as the work time based on the
detailed painting routes, augmented by penalties for constraint
violations.

Instead of directly computing the objective and constraint function
values from the upper-layer solution $\bm{x}^{(u)}$, the proposed method
calculates them based on the lower-layer subproblem's solutions
$\bm{x}^{(l)}$.
That is the proposed method calls a lower-layer optimizer to design
detailed routes and then computes these values accordingly.
In terms of GA, the detailed route $\bm{x}^{(l)}$ produced from the
genotype $\bm{x}^{(u)}$ serves as its phenotype.

\begin{table*}[t]
\caption{Constraints that must be considered for vehicle body painting.}
\label{table:constraints}
\centering
\begin{tabular}{@{}p{19mm}@{~~}l@{~~}p{90mm}@{~~~~}l@{~~}p{32mm}@{}}
\hline
Category & ID & Constraint & Strength & Handling technique\\
\hline
\hline
Robot arm 
\refstepcounter{const}\label{const:arm_area} & Constraint~\theconst 
         & Arm $a$ head must always remain within its operating range $\gamma_a$
           from its center $\bm{r}_{a, 0}$ for all $a \in \{1, \ldots, n_{arms}\}$.
         & Strong & Penalty function $p_{Constraint \ref{const:arm_area}}$,
                    \refstepcounter{repair}\label{repair:arm_area} 
                    Repair operator~\therepair \\ 
(spatial)
\refstepcounter{const}\label{const:arm_start_pos} & Constraint~\theconst 
         & Arm $a$ must return to its start position $\bm{r}_{a,0}$
           after finishing painting all assigned segments.
         & Strong & Lower-layer optimizer \\
\refstepcounter{const}\label{const:arm_body_col} & Constraint~\theconst 
         & Arm $a$ must not collide with the vehicle body.
         & Strong & Lower-layer optimizer\\
\refstepcounter{const}\label{const:arm_col} & Constraint~\theconst 
         & The distance between any tow arms $a_1$ and $a_2$ must always exceed $\gamma_{col}$ for all $a_1, a_2 \in \{1,\ldots, n_{arms}\}$ with $a_1 \neq a_2$.
         & Strong & Penalty function $p_{Constraint \ref{const:arm_col}}$\\
\hline
Robot arm
\refstepcounter{const}\label{const:work_time} & Constraint~\theconst 
         & Arm $a$ must finish painting all assigned segments within the specified time $t_p$, i.e., $t_a \leq t_p$.
         & Strong & Objective function\\
(temporal)
\refstepcounter{const}\label{const:arm_vel} & Constraint~\theconst 
         & Arm $a$ must paint at speed $v_{sp}$; when not painting, its
           speed must not exceed $v_{mv}$.
         & Strong & Lower-layer optimizer\\
\hline
Paint quality
\refstepcounter{const}\label{const:all_seg} & Constraint~\theconst 
         & Each paint path segment on a vehicle body must be painted
           exactly once.
         & Strong & Solution representation\\
\refstepcounter{const}\label{const:one_seg} & Constraint~\theconst 
         & Each paint path segment must be painted continuously from
           one end to the other, following its inherent shape.
         & Strong & Lower-layer optimizer\\
\refstepcounter{const}\label{const:btm_to_up} & Constraint~\theconst 
         & For a vertically mounted panel, painting must proceed
           sequentially from bottom to top, although a limited number of
           violations $\epsilon$ is acceptable.
         & Strong 
                  \refstepcounter{repair}\label{repair:btm_to_up} 
                  Repair operator~\therepair \\ 
\refstepcounter{const}\label{const:less_arm} & Constraint~\theconst 
         & Each panel should be painted using as few arms as possible.
         & Weak & \refstepcounter{repair}\label{repair:less_arm}
           Repair operator~\therepair \\ 
\refstepcounter{const}\label{const:bnd_height} & Constraint~\theconst 
         & The boundary heights of each arm's painting region should
           be uniformly aligned across the entire vehicle.
         & Weak & Initial solution generation\\
\hline
Equipment and
\refstepcounter{const}\label{const:line} & Constraint~\theconst 
         & A vehicle must be painted while moving in a production line
           at a constant speed.
         & Strong & Lower-layer optimizer\\
maintainability
\refstepcounter{const}\label{const:pri_fwd_arm} & Constraint~\theconst 
         & When multiple arms paint a single panel, painting should
           commence with the arm in the front row.
         & Strong & Repair operator~\ref{repair:btm_to_up} \\
\refstepcounter{const}\label{const:back_door} & Constraint~\theconst 
         & The last arm must not be assigned to paint the back door.
         & Strong & \refstepcounter{repair}\label{repair:back_door} 
           Repair operator~\therepair \\ 
\refstepcounter{const}\label{const:symmetric} & Constraint~\theconst 
         & A vehicle's painting route should be designed to be
           symmetric with respect to the left and right sides.
         & Weak & Lower-layer optimizer \\
\refstepcounter{const}\label{const:synchro} & Constraint~\theconst 
         & Among symmetrically arranged arms, the start times for
           painting must be synchronized whenever the painting panel
           changes.%
         & Weak & Lower-layer optimizer\\
\refstepcounter{const}\label{const:contami} & Constraint~\theconst 
         & Minimize paint contamination of other arms during the
           painting process.
         & Weak & Lower-layer optimizer \\     
    \hline
\end{tabular}
\end{table*}        

\subsubsection{Constraints}
\label{ssec:constraints}

Table~\ref{table:constraints} shows a list of the constraints
considered in the proposed method, which we categorize into three main
types: constraints on robot arms, on painting quality, and on
equipment and maintenability.
A solution is considered feasible if it satisfies all constraints
labeled as ``Strong,'' while meeting ``Weak'' constraints is highly
desirable.
Our method addresses these constraints through variable
representations, penalty functions in the upper-layer subproblem, repair
operators, and the lower-layer optimizer.

The constraints on robot arms split into spatial constraints and those
related to time or velocity.
Constraint \ref{const:arm_area} ensures that the robot arm's head
always lies within its operation range, meaning that the coordinate
$\bm{r}_{a,t}$ of arm $a$ at time $t$ (with $t \in \{1,\ldots, t_{max}\}$)
remains within the distance $\gamma_a$ from its movable center $\bm{r}_{a, 0}$.
\begin{equation}
 g_{Constraint~\ref{const:arm_area}, a, t}(\bm{x}^{(l)}) = \gamma_a - d(\bm{r}_{a,t},\bm{r}_{a, 0})  \geq 0
\end{equation}
where $d$ denotes the distance function.
The lower-layer optimizer evaluates whether solution $\bm{x}$
satisfies this constraint when generating a detailed painting path.
In cases where the target segment lies outside the arm's operating
range and painting cannot occur, three cases arise:
(a) the segment never enters the arm's motion range, (b) the segment
exits the working area during painting, and (c) the segment already
lies outside the range before painting begins.
The proposed method addresses case (a) with repair operator
\ref{repair:arm_area} and cases (b) and (c) with a penalty function.

Constraint \ref{const:arm_body_col}, which prevents contact between
arms and a vehicle body, factors into the lower-layer optimizer's design of 
detailed paths.
In vehicle painting, the arm head maintains a fixed distance from the
target surface to avoid contact and to maintain painting quality;
moreover, even if the arm risks colliding with the vehicle when moving
between the front hood, roof, or back door and the side, designing its
path to travel parallel to the painted panel until it reaches the next
panel's edge prevents such contact.

Constraint \ref{const:arm_col} on robot arm collisions guarantees that
the distance between any two distinct arms always remains at least
$\gamma_{col}$.
The lower-layer solver checks this condition during path generation,
and, after producing the detailed paths for all arms, it adds a penalty
to the objective function proportional to the time intervals during
which the inter-arm distance falls below the threshold.

Constraint \ref{const:work_time} requires each arm to complete
painting all designated surfaces and return to its starting point
within the prescribed time to prepare for the next vehicle.
Therefore, our method minimizes the maximum operating time among all
arms to satisfy this condition.

A key feature of our method lies in addressing painting quality
constraints.
Constraint~\ref{const:all_seg} guarantees that every painting path
segment on the vehicle receives exactly one coat and must be strictly
satisfied.
Therefore, when the lower-layer optimizer designs a painting path, any
unpainted segments trigger a penalty in the objective function
proportional to their count.

Constraint~\ref{const:one_seg} serves as one of the primary conditions
for ensuring painting quality.
In automobile painting, each panel is generally painted individually,
and the spraying process must cover the panel's width continuously at a
constant speed without any stops or interruptions.
This process involves painting a straight line, making a slight height
adjustment, and then continuing with another straight line until the
panel is entirely painted.
Thus, our method inputs painting path segments as described in
Section~\ref{ssec:overview} and ensures the arm refrains from any
non-painting actions, such as waiting, until it completes the current
segment.
In addition, the lower-layer optimizer generates the painting path
while maintaining the prescribed speed $v_{sp}$ for each vehicle.

Constraint~\ref{const:btm_to_up} prevents paint drips---i.e., the
downward movement of paint on sloped surfaces before it dries,
resulting in thick layers---thus enhancing painting quality.
For vertical body surfaces, the arm must paint sequentially from
bottom to top while moving horizontally back and forth, ensuring that
the order remains consistent on each panel.
Continuous bottom-to-top painting minimizes the visibility of spray
seams; however, due to installation constraints, the lower section of
the vehicle must be painted after the middle and upper sections.
Thus, Constraint~\ref{const:btm_to_up} allows at most a few
violations, denoted by $\epsilon$, per panel.
\begin{equation}
   g_{Constraint~\ref{const:btm_to_up}}(x) =  \epsilon - c\left(e(s_{b_m,l-1}) \geq e(s_{b_m,l})\right) \geq 0 
   \label{eq:const_order} \\
\end{equation}
where $c(\mathcal{C})$ represents the number of times condition $\mathcal{C}$ occurs.
We denote the painting locations on panel $b_m$ (with $m \in \{1, \ldots, n_{panels}$) by 
$s_{b_m,1}, s_{b_m, 2}, \ldots, s_{b_m, l}, \ldots$, where
a larger index $l$ indicates a higher position.
Let $e(s_{b_m, l})$ denote the time when the painting at path segment
$s_{b, l}$ occurs.
In principle, for adjacent segments 
$s_{b_m,l-1}$ and $s_{b_m,l}$, the painting time must satisfy
$e(s_{b_m,l-1}) < e(s_{b_m,l})$;
however, the proposed method allows upt to $\epsilon$ violations.
To produce solutions that meet this constraint in the upper-layer
optimizer, we incorporate repair operator \ref{repair:btm_to_up}
described in Section~\ref{sssec:repair} and employ a penalty function
to permit $\epsilon$ violations.

Constraint~\ref{const:less_arm} stands as one of the fundamental
conditions for maintaining painting quality. 
Considering both the painting quality and the operating time of the
robot arms in automotive processes, each panel should ideally be
painted by as few arms as possible. 
This approach minimizes the risk of seams, discrepancies in painting
time, and the complexity of the painting path.

Constraint~\ref{const:bnd_height} serves as another critical condition
for preserving painting quality. 
When painting vehicle side panels such as the front fender, front
door, rear door, and rear fender, one arm cannot manage an entire
panel alone, which forces multiple arms to divide the work. 
This division creates a slight delay between the one arm's finishing
time and the next arm's starting time, resulting in an undesirable
border in the painting area between the arms, although it is almost
invisible to the human eye.
Therefore, aligning the boundary heights between the painting areas on
neighboring panels improves overall quality and simplifies maintenance
by clearly identifying which arm handled each section.

In addition to constraints on the robot arms and painting quality, our
proposed method also considers equipment and maintenance
constraints. 
Constraint~\ref{const:arm_vel} requires that the vehicle body moves at
a constant speed, so the lower-layer optimizer generates detailed
painting paths that factor in this velocity.
Constraint~\ref{const:back_door} addresses the risk that when the last
arm paints the vehicle's 
back door, paint may splash onto other
arms or tools; 
any violation triggers repair operator \ref{repair:back_door}
described in Section~\ref{sssec:repair}.
Furthermore, Constraint~\ref{const:pri_fwd_arm} improves
maintainability by stipulating that when multiple arms paint a panel,
the arm at the front must start first. 
However, if we allow a violation of Constraint~\ref{const:btm_to_up},
we similarly permit a violation of
Constraint~\ref{const:pri_fwd_arm}, meaning that if the lower portion
of a panel is painted after the upper portion, the arm handling the
lower section may come from the front row.

Constraint~\ref{const:symmetric} mandates a left–right symmetric
design of the vehicle's painting path to enhance both painting quality
and maintainability. 
In our approach, the upper- and lower-layer optimizers first design
detailed trajectories for the arms on one side of the vehicle, and then
the lower-layer optimizer mirrors these routes to the other side.

Constraint~\ref{const:synchro}, which builds on
Constraint~\ref{const:symmetric}, acknowledges that even if the overall
path is symmetric, the lateral (z-axis) movements of a symmetric pair
of arms may not always align perfectly---especially under
Constraint~\ref{const:contami} discussed later. 
Therefore, the lower-layer solver generates detailed painting paths
that enable the corresponding pair of arms to start painting
simultaneously at the beginning of each panel.

Constraint~\ref{const:contami} specifies an exception to
Constraint~\ref{const:symmetric}. 
When the left and right arms follow symmetric paths to paint areas
such as the hood, roof, and back door, collisions may occur at the
vehicle's center. 
To prevent these collisions, we mirror the detailed trajectory from one
side to the other so that each arm pair paints in parallel while
maintaining equal spacing. 
However, relying solely on mirroring may keep the two arms to remain
at a constant distance for too long, 
which can cause cross-contamination of their respective arm heads.
Therefore, our method delays the start time on one side to avoid prolonged
fixed spacing between the corresponding arm heads.

\subsection{Algorithm of the upper-layer optimizer}

\begin{algorithm}[t]
  \caption{Upper-layer solver using GA}
  \label{fig:algorithm_overview}
  \begin{algorithmic}[1]
  \Require All paint path segments $\bm{s}$, 
           production equipment parameters
  \Ensure The best solution $\bm{x}^*$
  \State Create initial population
  \State Evaluate all individuals (solutions)
  \While {$g < n_{gen}$}
    \State Select parents
    \State Apply crossover operator
    \State Apply mutation operator
    \State Apply repair operators
    \State Evaluate all individuals (solutions)
    \State 
           Increment $g$ by 1
  \EndWhile
  \State \Return the best solution $\bm{x}^*$
  \end{algorithmic}
\end{algorithm}

Algorithm~\ref{fig:algorithm_overview} presents the details of our
GA-based upper-layer optimization algorithm.
Our upper-layer optimizer follows the standard GA procedure:
it first generates an initial population of solutions, then repeatedly
evaluates and regenerates candidate solutions until it reaches the
generation limit.
When regenerating candidates, it applies crossover (which produces two
new individuals from two parent solutions), mutation (which introduces
low-probability variations into individual candidates), and repair
operators to resolve constraint violations. 
In selecting parent solutions during crossover, we adopt the commonly
used tournament selection
scheme~\cite{mitchell1998introduction,filipovic2003fine}, whereby we
randomly choose $n_t$ individuals from the population and select the
one with the highest objective function value; here, $n_t$ denotes the
tournament size.

For the parent individuals selected by tournament selection, our
method applies crossover and mutation operations to generate
offspring. 
We use Order Crossover---a technique commonly used in GA for TSP and
VRP~\cite{davis1985applying,deep2011new}---which preserves the
visitation order of specific painting path segments from one parent.
Specifically, one parent contributes a subset of genes while the other
provides the remaining genes in their relative order. 
Thus, the offspring
$\bm{x}_{c1}^{(u)} = \{x_{c1, k}\}_{k \in \{1, \ldots, n_{dim}\}}$
derives from the parent individuals
$\bm{x}_{p1}^{(u)} =\{x_{p1, k}\}_{k \in \{1, \ldots, n_{dim}\}}$ and
$\bm{x}_{p2}^{(u)} =\{x_{p1, k}\}_{k \in \{1, \ldots, n_{dim}\}}$ as follows:
\begin{equation}
x_{c1, k} = 
\left\{ 
\begin{array}{ll}
x_{p1, k} & \mbox{if}~ k_s \leq k \leq k_e \\
x'_{p2,r(k)} & \mbox{otherwise}
\end{array}
\right.
\end{equation}
where $k_s$ and $k_e$ denote the positions at which crossover occurs,
$\bm{x}'^{(u)}_{p2} = \{ x_{p2, 1}', \ldots, x_{p2, n_{dim} - (k_e - k_s +
  1)}'\}$ denotes a sequence derived from $\bm{x}_{p2}^{(u)}$ after
excluding the painting segments that lie within the interval $(k_s,
k_t)$ in $\bm{x}_{p1}^{(u)}$, and $r(k)$ is a function that returns
the index in $\bm{x}'^{(u)}_{p2}$ corresponding to position $k$.

Furthermore, after crossover, the upper-layer optimizer uses an
inversion operation~\cite{lee2003improved} as the mutation operator,
which swaps two randomly chosen elements in a candidate
solution with a certain probability. 
Next, it applies repair operators to the offspring generated by the
crossover and mutation steps to address constraint violations. 
Our proposed method curbs these violations by implementing four types
of repair operations, as detailed in Section~\ref{sssec:repair}.

\subsection{Mechanisms for constraint handling}

Our proposed method addresses the various constraints listed in
Table~\ref{table:constraints} by refining the representation of
candidate solutions, applying repair operations when violations occur,
imposing penalties, modifying a population initialization process, and
so on.
The column of ``handling methods'' in Table~\ref{table:constraints}
illustrates how our method handles each constraint.
By employing the expressions described in
Section~\ref{sssec:variables} and the repair operations outlined in
Section~\ref{sssec:repair}, our method ensures that
Constraint~\ref{const:all_seg} is always satisfied.
Moreover, minimizing the
objective function $f(\cdot)$ yields solutions that satisfy
Constraint~\ref{const:work_time}. 
Our method employs penalty functions to resolve violations
of Constraints~\ref{const:arm_area} and \ref{const:arm_col}.
In addition, it accounts for Constraints \ref{const:arm_start_pos},
\ref{const:arm_body_col}, \ref{const:arm_vel}, \ref{const:line},
\ref{const:symmetric}, \ref{const:synchro}, \and ref{const:contami}
during the detailed path planning in the lower-layer optimizer to
generate compliant solutions.

\subsubsection{Penalty functions}
\label{sssec:penalty}

To resolve violations of Constraints~\ref{const:arm_area} and \ref{const:arm_col},
the proposed method employs penalty functions. 
For Constraint~\ref{const:arm_area}---which concerns the robot arm's
operating range while the target vehicle moves at a constant speed---
it addresses violations through both penalty functions and repair
operators.
Specifically, our penalty function $p_{Constraint \ref{const:arm_area}}$
calculates a penalty based on the total number of unpainted sites
$n_{unvisits}$ resulting from the vehicle moving out of arm $a$'s
range, and the duration $t_{out}$ during which a segment is painted
while the arm is outside its operating range.
Because our lower-layer optimizer follows a simple greedy algorithm,
it continues to paint even if the arm operates outside its designated
range; the time spent in such an infeasible state is recorded as
$t_{out}$.
\begin{equation}
  p_{Constraint \ref{const:arm_area}}  =  
  \sum_a  \rho_{out} \times t_{out}
  (\bm{r}_a)
  + \rho_{unvisits} \times n_{unvisits}
  (\bm{r}_a)
  \label{eq:rom_unvisit_penalty}\\
\end{equation}
where $\rho_{out}$ and $\rho_{unvisits}$ denote penalty values, and $\bm{r}_a = \{\bm{r}_{a,t}\}_{t \in {1, \ldots, t_{max}}}$ .

Constraint~\ref{const:arm_col}, which governs collisions between robot
arms, applies to environments where multiple arms must maintain sufficient
spacing. 
Such violations occur infrequently; however, when a collision is
detected, our method penalizes it by adding $p_{Constraint \ref{const:arm_col}}$ to the
objective function:
\begin{equation}
  p_{Constraint \ref{const:arm_col}} =  \rho_{col} \times t_{col}
  \label{eq:col_penalty}
\end{equation}
where $t_{col}$ denotes the time that collisions occur and $\rho_{col}$
denotes a penalty value.

\begin{figure}[t]
  \centering
  \begin{tabular}{cc}
    \includegraphics[width=0.23\textwidth]{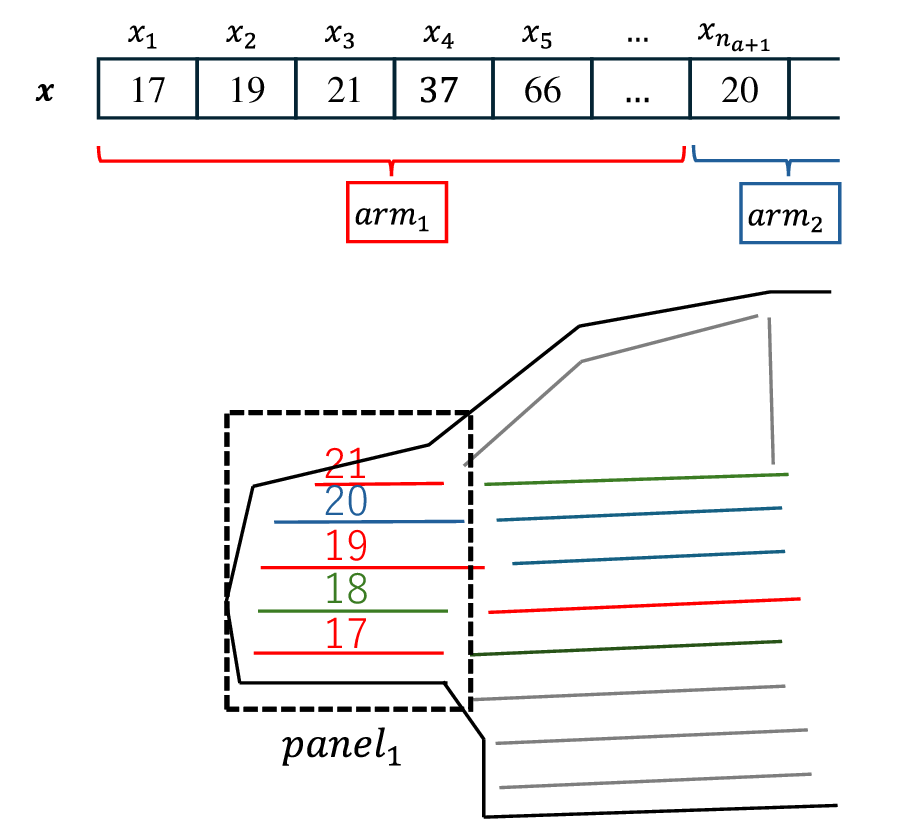} &
    \includegraphics[width=0.23\textwidth]{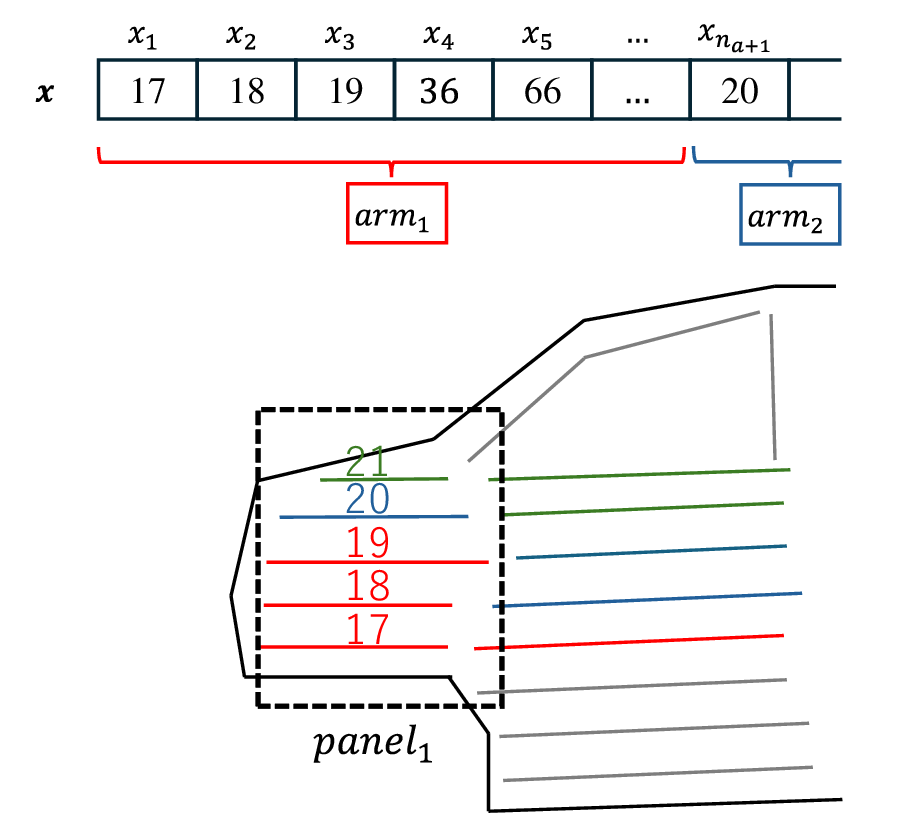} \\
    {\footnotesize (a) Before} &
    {\footnotesize (b) After} \\
  \end{tabular}
  \caption{Example of applying Repair operator~\ref{repair:btm_to_up}.}
  \label{fig:continuity_repair}
\end{figure}

\subsubsection{Repair operators}
\label{sssec:repair}

To resolve violations for Constraints \ref{const:arm_area},
\ref{const:btm_to_up}, \ref{const:less_arm},
\ref{const:pri_fwd_arm}, and\ref{const:back_door},
our method applies repair operators
specifically designed for each constraint.
In particular, for Constraint~\ref{const:arm_area}, it obtains a fully
compliant solution by combining the penalty function described in
Section~\ref{sssec:penalty} with Repair operator~\ref{repair:arm_area}.
In the upper-layer optimizer, violations of
Constraint~\ref{const:arm_area} occur when robot arm $a$ receives
painting segments that it cannot ever reach---i.e., segments that
remain outside its operating range regardless of operating time.
When such a violation occurs, Repair operator~\ref{repair:arm_area}
swaps the problematic segment with one assigned to another arm; it
scans the upper-layer solution from the beginning and performs the
exchange as soon as it finds a segment whose swap eliminates the
violation.

Upon confirming a breach of the painting order requirement specified
in Constraint~\ref{const:btm_to_up} (Eq.~(\ref{eq:const_order})),
our method deploys Repair operator~\ref{repair:btm_to_up} to
concurrently modify the segment assignments and visiting order,
ensuring that each robot arm paints a continuous region.
For each panel $b_m$ on the vehicle's side and back door, our method begins by
rearranging the segments for each arm so that they remain continuous.
In detail, it first determines the number of segments
$|\bm{s}_{\bm{x}_a, b_m}|$ that each arm $a$ should paint within panel
$b_m$.
Next, it reassigns the segments 
$s_{b_m,1}, \ldots, s_{b_m, |\bm{s}_{b_m}|}$
(where lower indices denote segments at the bottom) in sequence from
the frontmost arm, based on $|\bm{s}_{\bm{x}_a, b_m}|$.
Consequently, Repair operator~\ref{repair:btm_to_up} addresses
violations of both Constraints~\ref{const:btm_to_up} and
\ref{const:pri_fwd_arm}.

Figure~\ref{fig:continuity_repair} illustrates an example where our
method applies Repair operator~\ref{repair:btm_to_up} to three arms
(arm 1, 2, and 3) painting panel 1. 
Before the repair, arm 1 handled segments 17, 19, and 21, arm 2 was
assigned segment 20, and arm 3 received segment 18, resulting in
fragmented, non-contiguous routes.
Repair operator~\ref{repair:btm_to_up} then reassigns segments so that
arm 1 covers segments 17, 18, and 19, arm 2 handles segment 20, and
arm 3 takes segment 21, thereby maintaining each arm's allocated
count while ensuring a continuous painting sequence from the bottom
upward.

Repair operator~\ref{repair:less_arm} ensures compliance with
Constraint~\ref{const:less_arm} by swapping the assignments of
painting path segments between two arms that paint two different
panels.
First, our method focuses on two panels, $b_{m_1}$ and $b_{m_2}$ (with $m_1, m_2
\in \{1, \ldots, n_{panels}\}$ and $m_1 \neq m_2$), and two arms, $a_1$
and $a_2$ (with $a_1, a_2 \in \{1, \ldots, n_{arms}\}$ and $a_1 \neq a_2$),
that paint segments on panels $b_{m_1}$ and $b_{m_2}$.
Here, we denote a range of assigned segments in the upper-layer
solution $\bm{x}_i$ for arm $a$ as $\bm{q}_a$, i.e., $q_a = \{(a-1)
\times n_{arms} + 1, \ldots a \times n_{arms} \}$.
Repair operator~\ref{repair:less_arm} then exchanges the assignments
of panels $b_{m_1}$ and $b_{m_2}$ between arms $a_1$ and $a_2$ to satisfy
Constraint~\ref{const:less_arm}, i.e., 
\begin{equation}
 x'_{i, k} = \left\{ \begin{array}{ll}
              x_{i, r'(k, a_2, b_{m_2})} & \mbox{if}~ k \in \bm{q}_{a_1} \land~ x_{i, k} \in \bm{s}_{b_m1} \\
              x_{i, r'(k, a_1, b_{m_1})} & \mbox{if}~ k \in \bm{q}_{a_2} \land~ x_{i, k} \in \bm{s}_{b_m2} \\
              x_{i, k} & \mbox{otherwise}
              \end{array} \right.
\end{equation}
where $r'(k, a, b_j)$ returns
the index (gene position) within $\bm{x}_i$ that contains the $k$-th
segment assigned to arm $a$ and belonging to panel $b_j$.

Repair operator~\ref{repair:back_door} sequentially swaps the back
door painting segments assigned to the last arm $a_{n_{arms} / 2}$
with segments from non-back door panels, ensuring that no
Constraint~\ref{const:arm_area} violation occurs.

  \begin{algorithm}[t]
    \caption{Initial popluation creation to align paint boundary heights.}
    \label{fig:horizontal_init_algorithm}
    \begin{algorithmic}[1]
    \Require All paint path segments $\bm{s}$,              
             reference panel $b_{ref}$ 
    \Ensure Initial population $\bm{p}_{init}$
    \State Let $\bm{h}$ be a set of heights $h_1, \ldots, h_{n_{arms}/ 2 - 1}$
            representing the boundaries between every
            $2|\bm{s}_{b_{ref}}| / n_{arms}$ consecutive segments
    \State Create the first initial solution $\bm{x}^{(init)}_1$
           according to $\bm{h}$
    \State $\bm{p}_{init} \leftarrow \bm{p}_{init} \cup \bm{x}^{(init)}_1$
    \State $\bm{p}_{init} \leftarrow \bm{p}_{init} \cup \Call{ extend }{\bm{p}_{init}, \bm{h}, \{\}, 1}$
    \State \Return $\bm{p}_{init}$
    \Statex
      \Function { extend }{$\bm{p}_{init}, \bm{h}, \bar{\bm{h}} (= \{\bar{h}_1, \ldots, \bar{h}_{i-1} \}), i$}
      \If{$|\bm{p}_{init}| = n_{pop} - 1$}
        \State \Return $\bm{p}_{init}$
      \ElsIf{$i = n_{arms}$}
        \State Create initial solution $\bm{x}^{(init)}_j$  according to $\bm{h}$
        \State $\bm{p}_{init} \leftarrow \bm{p}_{init} \cup \bm{x}^{(init)}_j$
      \EndIf
      \For{$j \in \{1, \ldots, \delta\}$}
        \State Let $\bar{h}_i$ be the height of the boundary obtained 
        \Statex ~~~~~~~~~by raising the $i$-th boundary in $\bm{h}$ by $j$ segments
        \State $\Call{ create\_init\_pop}{\bm{h}, \bar{\bm{h}} \cup \bar{h}_i,  i+1}$
        \State Let $\bar{h}_i$ be the height of the boundary obtained 
        \Statex ~~~~~~~~~by lowering the $i$-th boundary in $\bm{h}$ by $j$ segments
        \State $\Call{ create\_init\_pop}{\bm{h}, \bar{\bm{h}} \cup \bar{h}_i,  i+1}$
      \EndFor
      \State \Return $\bm{p}_{init}$
      \EndFunction
    \end{algorithmic}
 \end{algorithm}

 \begin{algorithm}[t]
    \caption{Lower-layer solver}
    \label{alg:simulator}
    \begin{algorithmic}[1]
    \Require A solution of the upper-layer subproblem $\bm{x}^{(u)}$, 
             all paint path segments $\bm{s}$,
             production equipment parameters 
    \Ensure A solution of the lower-layer subproblem $\bm{x}^{(l)}$
    \State Find the first segment $x_k$ (with $k=(a-1)\times n_{arms} + 1$)
           assigned to arm $a$. If $x_k$ includes a dummy segment,
           increment $k$ until $x_k$ does not include dummy
    \For{each time unit $t \in \{1, \ldots, t_{max}\}$}
      \For{each arm $a \in \{1, \ldots, n_{arms} / 2 \}$}
      \If{arm $a$ hs finished painting all its assigned ~~~~ \hspace*{6ex} segments}
        \State Move to its initial position and wait
      \ElsIf{
        $a$ is currently painting segment
        $x_k$
      }
        \State Paint
               $x_k$
               with velocity $v_{sp}$
        \If{
          $a$ finishes painting
          $x_k$
        }
          \State Increment $k$ until $x_k$ is not dummy
        \EndIf
      \Else
        \If{
          $a$ is at a panel border where a head angle  ~~~~~ \hspace*{9.2ex} change is neededarm head angle
        }
          \State Wait until the arm head angle adjusts
        \ElsIf{
            the entire area of segment $x_k$ is within ~~~~~ \hspace*{9.2ex} the arm $a$'s range
        }
          \If{
            $a$ is at the start position for
            $x_k$
          }
            \State Start painting 
                   $x_k$
          \Else
            \State Move to
                   $x_k$
                   with velocity $v_{mv}$
          \EndIf
        \Else
          \State Wait 
        \EndIf
      \EndIf
    \EndFor
    \State Increment $t$ by 1 
  \EndFor
  \State Copy the paths of arms $\{1, \ldots, n_{arms}/ 2\}$ to those of the corresponding arms $\{n_{arms}/2+1, \ldots, n_{arms}\}$
  \State Modify the copied paths so that the arms move in parallel on panels such
         as the hood, roof, and back door, where symmetric painting causes collisions
  \State Insert wait actions as necessary to synchronize the
         corresponding left and right arms to satisfy Constraint~\ref{const:synchro}
  \State \Return $\bm{x}^{(l)} (= \{\bm{r}_{a, t} \}_{a \in \{1, \ldots, n_{arms}\}, t \in \{1, \ldots, t_{max}\}})$
  \end{algorithmic}
  \end{algorithm}

\subsubsection{Initial population generation}
\label{ssec:init_pop}

As illustrated by Constraint~\ref{const:bnd_height} in
Table~\ref{table:constraints}, when engineers design the route, they
align the boundary heights of the painting regions assigned to each
arm across panels such as fenders and doors to ensure high painting
quality and ease of management. 
However, because vehicle models and production lines impose varying
conditions and additional constraints, establishing a strict
standard for this constraint proves challenging. 
Therefore, our method attempts to satisfy
Constraint~\ref{const:bnd_height} by employing an initial solution
with a horizontally aligned path.

Algorithm~\ref{fig:horizontal_init_algorithm} illustrates the procedure
for generating an initial solution with horizontally aligned features.
First, the upper-layer optimizer selects the panel on the vehicle
body's side that contains the maximum number of horizontally extended
painting path segments as the reference panel 
$b_{ref}$ 
and denotes the
number of its painting segments as $| \bm{s}_{b_{ref}} |$.
After setting the reference panel, it partitions the segments so that
$n_{arms}/2$ arms on one side each paint 
$\frac{2 | \bm{s}_{b_{ref}}  | }{ n_{arms}}$ 
segments.
The set of boundaries on the reference panel, denoted by $\bm{h}$,
contains heights of $n_{arms} / 2 - 1$ boundaries established between
every group of $\frac{2 | \bm{s}_{b_{ref}} |}{n_{arms}}$ consecutive
segments.

Based on the boundaries defined in $\bm{h}$, the upper-layer optimizer
determines the painting range for each arm.
Considering Constraints~\ref{const:btm_to_up} and \ref{const:pri_fwd_arm},
it allocates the lower region of the reference panel to the leading
arm and the upper region to the trailing arm.
After setting the painting boundaries on the reference panel, it replicates
the boundaries at the same height on the other panels to assign the painting
areas for each arm, thereby generating the first initial solution.

Next, the upper-layer optimizer generates candidate assignments by
shifting each of the $n_{arms}-1$ boundaries in $\bm{h}$ by an offset
ranging from 1 to $\delta$ segments.
After it determines the painting boundaries on the reference panel, it
sets corresponding boundaries at the same heights on the other panels
and adds these as initial solutions to the population. 
If these solutions alone fail to meet the predefined population size $n_{pop}$,
the optimizer supplements them with randomly generated individuals.

For vehicles where the painted area encompasses the roof, we merge the
roof with side panels (including doors and fenders) that are aligned
along the x-axis, and we treat both as reference panels.
In doing so, the optimizer interprets the roof's width as continuous
with the side panels' height, thereby segmenting the painting range
for the arms.
In addition, when a vehicle includes roof painting, it manages the hood
in the same manner by aligning their respective boundaries.

\begin{table*}[tb]
  \caption{Configurations for each vehicle model.}
  \label{table:experiment_setting}
  \centering
  \begin{tabular}{llll}
    \hline
      & $V_{1}$ & $V_{2}$ & $V_{3}$\\
    \hline\hline
    Number of painting path segments $n_{segs}$ & 268 & 260 & 218 \\
    Number of arms $n_{arms}$ & 4(8) & 4(8) & 3(6)\\
    Arm placement  & Fig.~\ref{fig:layout}(a) & Fig.~\ref{fig:layout}(a) & Fig.~\ref{fig:layout}(b) \\
    Radius of arm operation range $\gamma_{a}[\rm{mm}]$ & 2,800 & 2,800 & 2,800 \\
    Arm collision avoidance distance $\gamma_{col}[\rm{mm}]$ & 300 & 300 & 300 \\
    Arm velocity during painting $v_{sp}[\rm{mm} / \rm{s}]$ & 1,250 or 1,350 & 1,250 or 1,350 & 900 or 1,000 \\
    Arm velocity when not painting $v_{mv}[\rm{mm} / \rm{s}]$ & 1,250 & 1,250 & 900 \\
    Panels to be painted & Hood, door, fender, back door & Hood, door, fender, back door & Hood, roof, door, fender, back door\\
    Constraints to consider & All & All & All except Constraint~\ref{const:back_door} \\
    Velocity of production line $[\rm{mm} / \rm{s}]$ & 147 & 147 & 98 \\
    Prescribed time $t_p$ $[\rm{s}]$ & 47.5 & 47.5 & 66.0 \\    
    Number of dummy segments $n_d$ & 68 & 68 & 58 \\
    Allowed violations for Constraint~\ref{const:btm_to_up} $\epsilon$ & 1 & 1 & 0 \\
    Range for changing boundary hight $\delta$ & 3 & 3 & 5 \\
    \hline
  \end{tabular}
\end{table*}

\begin{figure*}[t]
  \centering
      {\small
  \begin{tabular}{@{}l@{~~}c@{\hspace{5mm}}c@{}}
    \rotatebox{90}{~~~~~~~~~~~~Top view} &
    \includegraphics[height=0.17\textheight,width=0.4\textwidth]{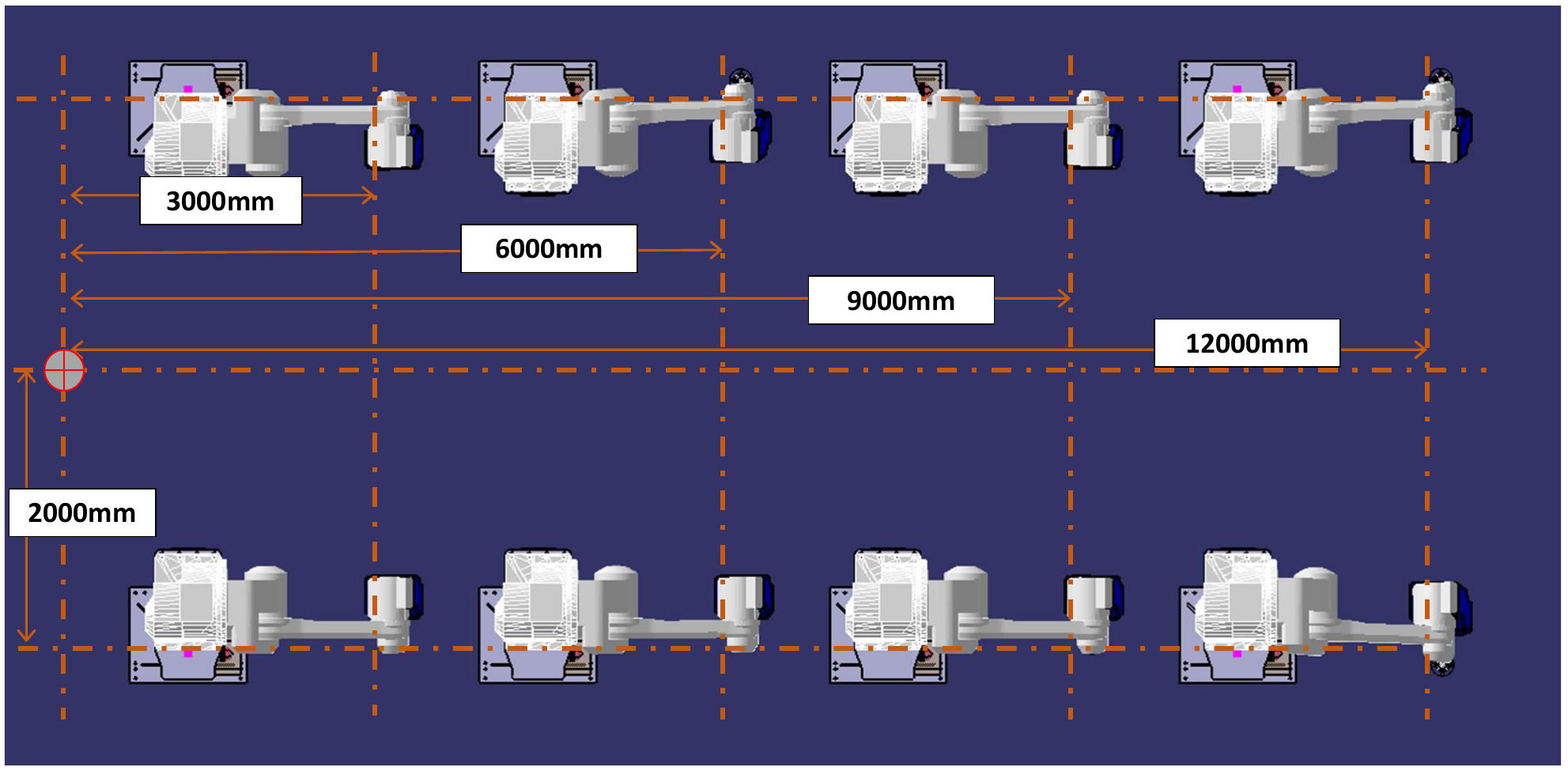} &
    \includegraphics[height=0.17\textheight,width=0.4\textwidth]{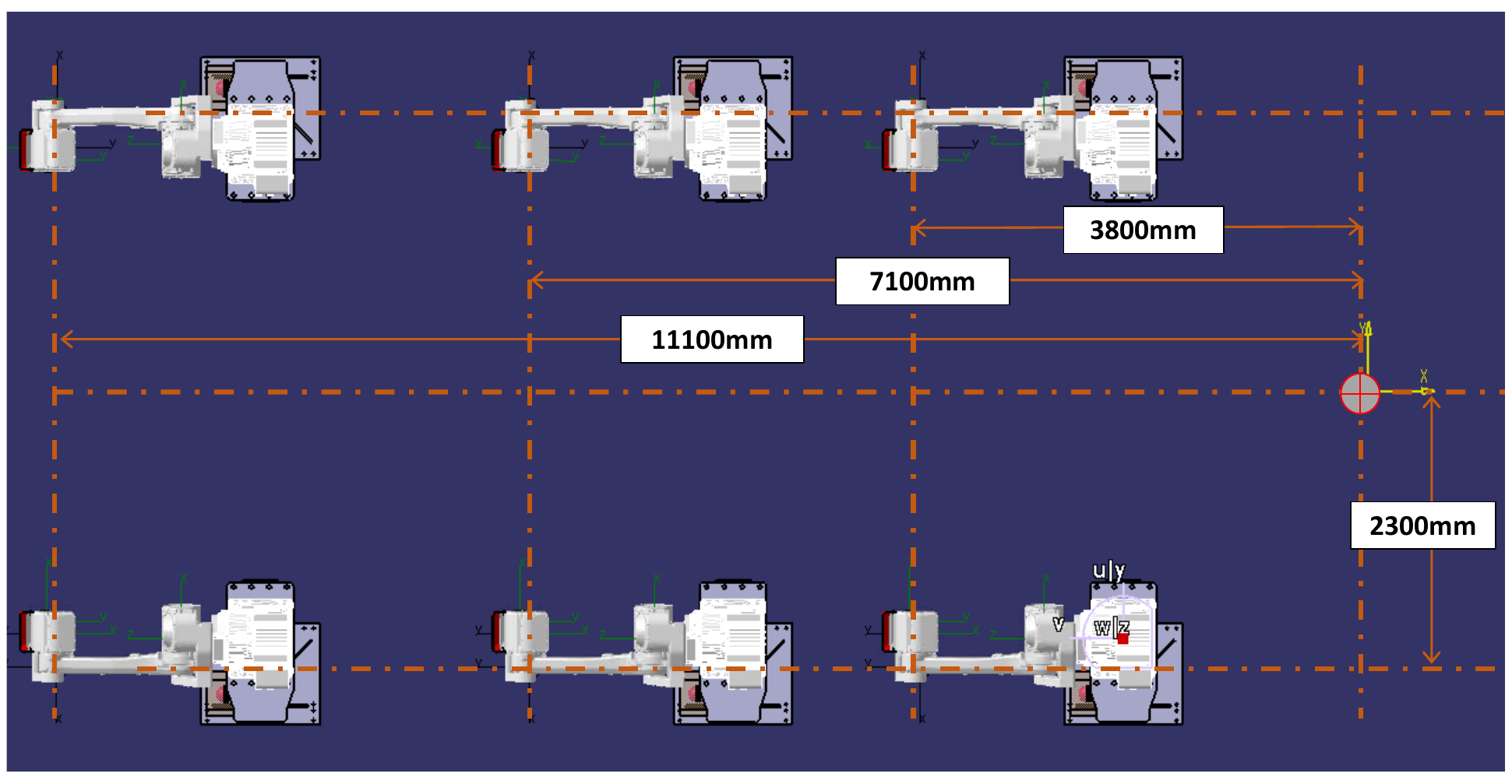}\\
    \rotatebox{90}{~~~~~~~~~~~~Side view} &
    \includegraphics[height=0.15\textheight,width=0.4\textwidth]{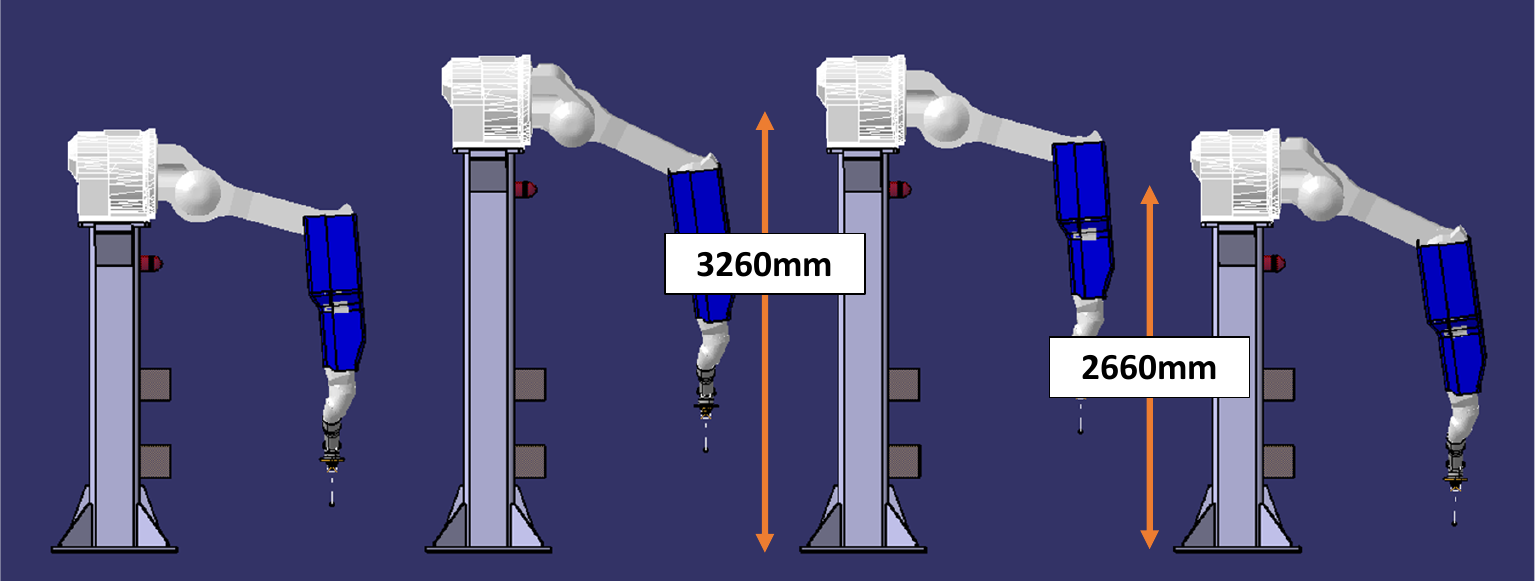} &
    \includegraphics[height=0.15\textheight,width=0.4\textwidth]{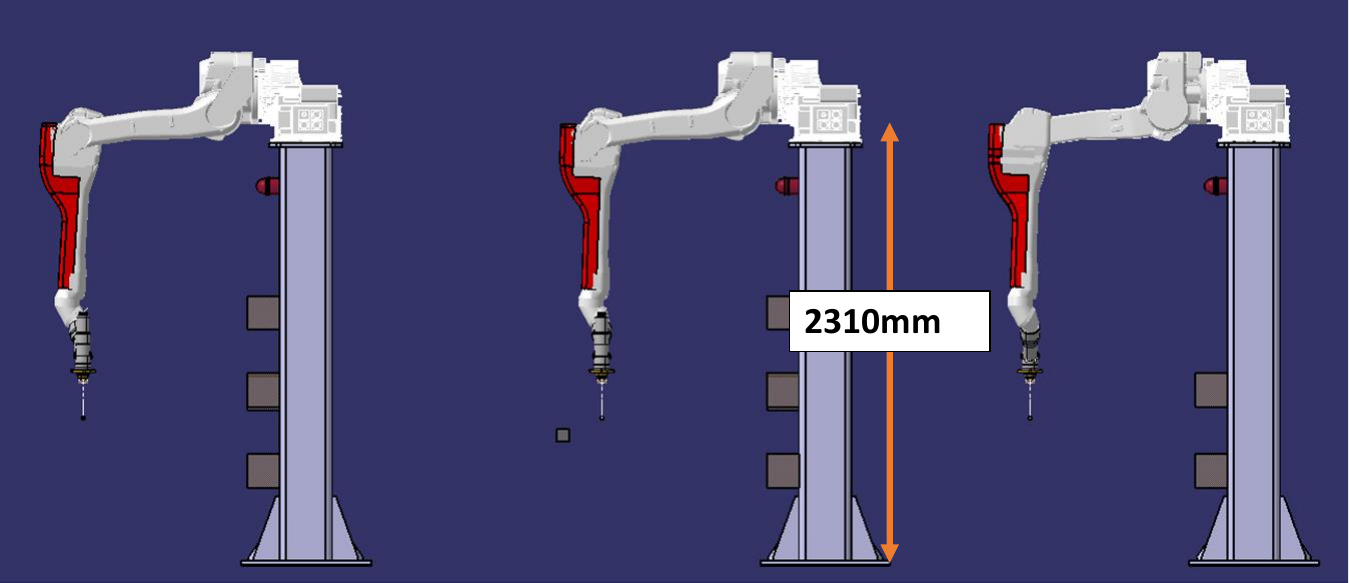} \\
     &(a) $V_{1},V_{2}$ produced  at $F_1$ & (b) $V_{3}$ produced at $F_2$: side view
  \end{tabular}
      }
  \caption{Arm arrangements in $F_1$ and $F_2$.}
  \label{fig:layout}
\end{figure*}

\subsection{Lower-layer optimizer}

Because the upper-layer subproblem involves assigning painting path
segments to robot arms and determining the sequence in which each area
is painted, the lower-layer solver employs a greedy approach to
determine the detailed trajectories of the arm heads based on the
solution of the upper-layer subproblem.
In the vehicle painting factories targeted by this study, three or
four robot arms are arranged on each side of a production line, with
painting regions divided between the left and right arms at a vehicle
body's center.
Additionally, due to Constraint~\ref{const:symmetric}, the painting paths must be
designed to be essentially symmetrical.
Consequently, the lower-layer optimizer of the proposed method first
generates the trajectories of the arm heads for one side and then
extends them to the other side while considering
Constraints~\ref{const:synchro} and \ref{const:arm_col}.
In this paper, we define a three-dimensional space by assigning the
x-axis to a vehicle's front-to-back direction, the y-axis to its
vertical direction, and the z-axis to its left-right direction.

\subsubsection{Constructing one-sided route}

Algorithm~\ref{alg:simulator} presents the lower-layer optimizer's
detailed algorithm.
The lower-layer optimizer computes the coordinates of all arms at each
time unit and checks for violations of
Constraints~\ref{const:arm_area} through \ref{const:btm_to_up} and
\ref{const:symmetric} through \ref{const:contami}.
By repeating this process, it derives the detailed painting routes for
all arms assigned to paint one vehicle body.
Each robot arm establishes its detailed path by sequentially visiting
its assigned painting segments.
Unlike a conventional VRP, these segments have a line-like shape, so the
arm must decide which endpoint to approach.
Furthermore, the target vehicle moves at a constant speed. 
A robot arm remains stationary until both endpoints of a painting
segment fall within its operating range, at which point it begins
moving. 
Upon reaching the segment, the arm travels from one endpoint to the
opposite at speed $v_{sp}$ while spraying paint.
When it arrives at the endpoint, it marks the segment as completed and
moves to the next segment, choosing the endpoint that minimizes travel
distance.
Additionally, when the arm transitions between panels with significant
orientation changes (e.g., from a side panel to the roof), it waits
for a preset duration to account for reorientation time.

While designing detailed routes in the lower-layer optimizer, it
computes metrics to detect constraint violations and to apply
penalties.
Specifically, it records $t_{out}$ (the time arm $a$ worked outside
its operating range), $n_{unvisits}$ (the number of segments
assigned but not painted), $t_{col}$ (the duration during which the
distance between two arm heads falls below $\gamma_{col}$), and the
painting start time for each segment.
These start times are used to assess violations of
Constraint~\ref{const:btm_to_up}.
Finally, the lower-layer optimizer returns the final operating time
$t_a$ of all arms to the upper-layer optimizer along with the above
metrics.

\subsubsection{Expansion to both sides}

The lower-layer optimizer creates detailed routes for both sides of
the vehicle body by designing the opposite side's painting path based
on the detailed route produced for one side.
Our method predefines three bilateral expansion rules for each panel
of the target vehicle:
{\renewcommand{\labelenumi}{(\arabic{enumi})}
\begin{enumerate}
\item Left–right symmetry,
\item Parallel movement, and
\item Parallel movement with unilateral delay.
\end{enumerate}}
Rule (1) requires that a pair of arms arranged symmetrically on the
vehicle perform mirror-image actions (flipped along the z-axis). 
Rules (2) and (3) apply to panels such as the hood, back door, and
roof---areas where symmetric routes might lead to collisions. 
In Rule (2), the arms paint in parallel while maintaining a constant
distance to avoid collisions; 
in Rule (3), one arm's movement is delayed to prevent the arms from
remaining at a fixed distance and continuously overlapping paint.

Based on these rules, when the lower-layer optimizer expands the route
from one side to the other, vehicle asymmetries and Rule (3) can cause
even corresponding arms to behave non-symmetrically. 
Therefore, we synchronize symmetric robot arms by aligning the start
times for painting new panels during transitions
(Constraint~\ref{const:synchro}), enabling the design of bilateral
routes that avoid collisions without significantly affecting other
routes.

\section{Evaluation}
\label{sec:experiments}

\subsection{Experimental setup}
\label{ssec:setup}

To validate our proposed method, experiments using three different
types of vehicle bodies, $V_1$, $V_2$, and $V_3$ produced at two two
factories, $F_1$ and $F_2$.
Table~\ref{table:experiment_setting} summarizes the experimental
conditions for these three vehicle types, and Fig.~\ref{fig:layout}
shows the arm placements in these two factories.
The Production environments varied considerably: Factory $F_1$ manufactures
vehicles $V_1$ (a multi-purpose vehicle) and $V_2$ (a commercial van),
while $F_2$ produces $V_3$ (an off-road sport utility vehicle).
At $F_1$, the production line featured eight arms---four on each side,
as shown in Fig.~\ref{fig:layout}(a)--- which paint the hood, side
panels (including fenders and doors), and back door.
In contrast, $F_2$'s line had six arms---three on each
side, as shown in Fig.\ref{fig:layout}(b)--- which paint the hood,
side panels, back door, and roof.
Since $V_3$'s arm configuration did not consider
Constraint~\ref{const:back_door}, we designed its route without that
constraint.
In this experiment, the arm's installation positions, operating
ranges, and movement speeds $v_{sp}$ and $v_{mv}$ are determined based
on actual production settings, and we assumed that the arm heads
always faced a body surface perpendicularly.

The proposed method was configured as follows: GA served as the
upper-layer optimizer with a population size $n_{pop}$ of 400 individuals, a 2\%
mutation rate, and a maximum number of generations $n_{gen} = 150$.
Penalty values for Constraints~\ref{const:arm_area} and
\ref{const:arm_col} were set to values that exceed the objective
function's value when no violations occur, i.e.,
$\rho_{out}=5.0\times10^2$, $\rho_{unvisits}=10^4$, and $\rho_{col}=10^3$.
The time unit $\mu$ in the lower-layer optimizer was set to $10^{-2}$
second.
Based on the setup outlined above, we ran five individual trials
per vehicle on a computer equipped with an AMD EPYC 7773X Processor
(64 cores, 2.2--3.5GHz).

\subsection{Experimental results}
\label{ssec:results}

\begin{figure*}[t]
  \centering
      {\small 
  \begin{tabular}{@{}c@{}c@{}c@{}}
    \vspace*{-6mm}
    \includegraphics[height=0.16\textheight,width=0.33\textwidth,trim={70mm 0mm 70mm 10mm},clip]{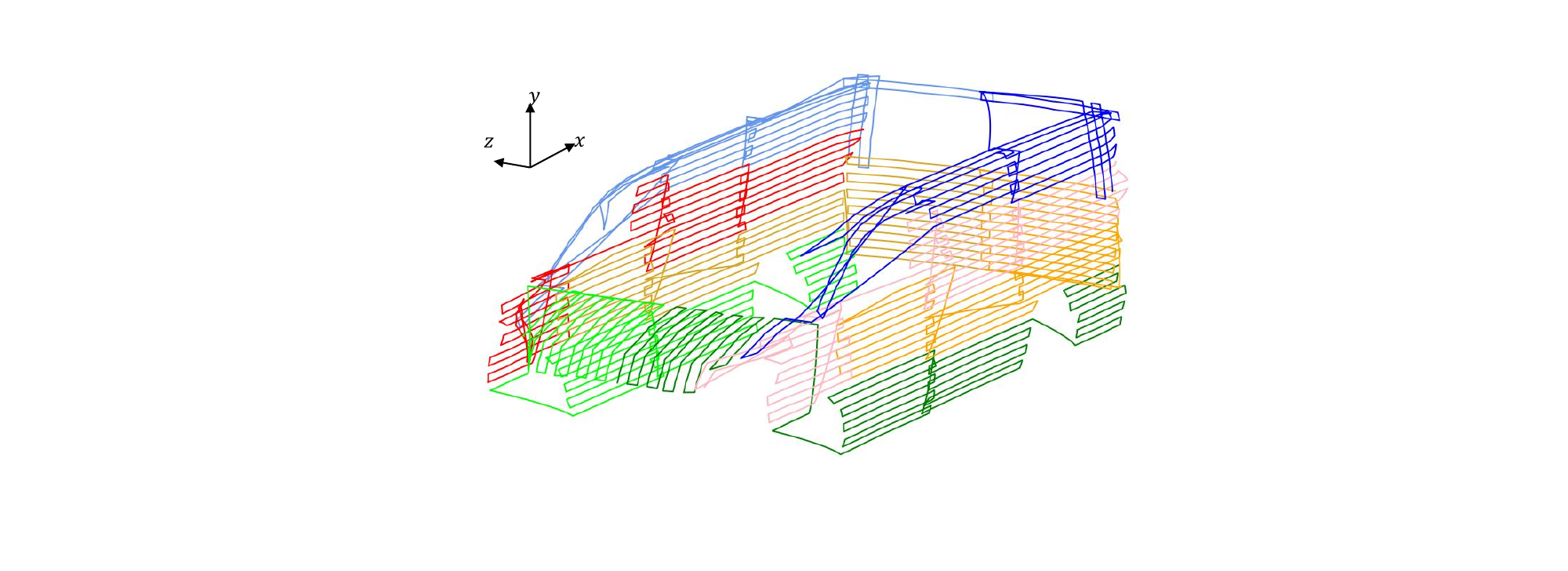} &
    \includegraphics[height=0.16\textheight,width=0.33\textwidth,trim={70mm 0mm 70mm 10mm},clip]{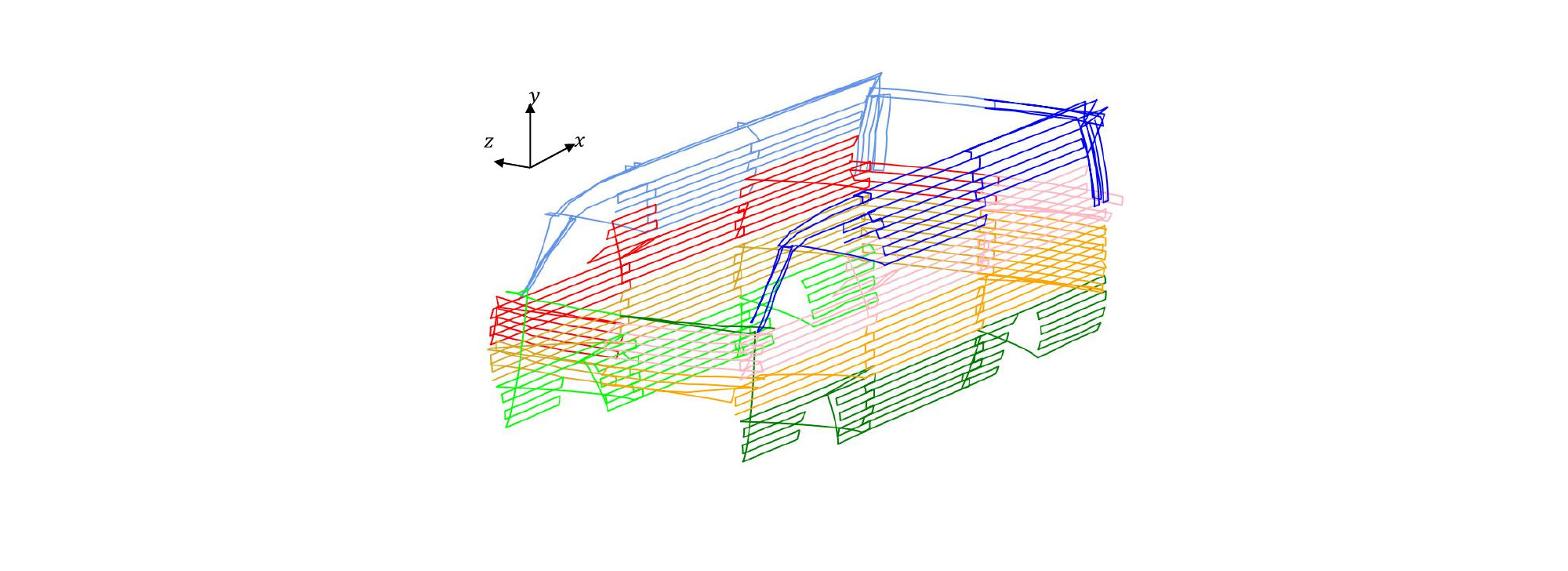} &
    \includegraphics[height=0.16\textheight,width=0.33\textwidth,trim={70mm 0mm 70mm 10mm},clip]{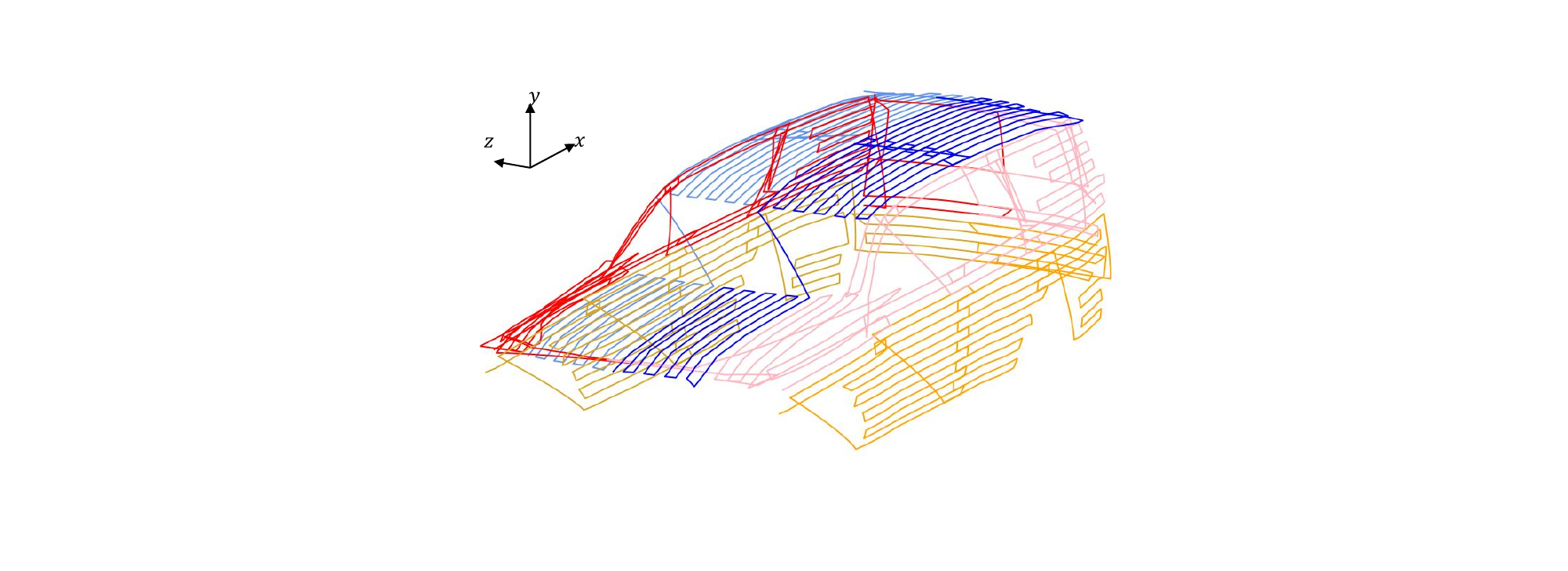}
    \\
    (a) $V_{1}$ & (b) $V_{2}$  & (c) $V_{3}$
  \end{tabular}
      }
  \caption{Example designed routes by the proposed method.}
  \label{fig:optimization_route}
\end{figure*}

Fig.~\ref{fig:optimization_route} shows the detailed painting paths
obtained by applying our method to the three vehicle types, 
with each arm's painting path distinguished by different colors.
The results confirmed that the proposed method could design routes
satisfying all constraints even when the number of painting segments
and arms varied across vehicle types.
Moreover, the routes enabled each arm to paint a continuous region,
thanks to the introduction of Repair operators~\ref{repair:btm_to_up}
and \ref{repair:less_arm} for addressing
Constraint~\ref{const:btm_to_up}.

\begin{figure*}[t]
  \centering
      {\small 
  \begin{tabular}{@{}c@{}c@{}c@{}}
    \includegraphics[height=0.2\textheight,width=0.33\textwidth]{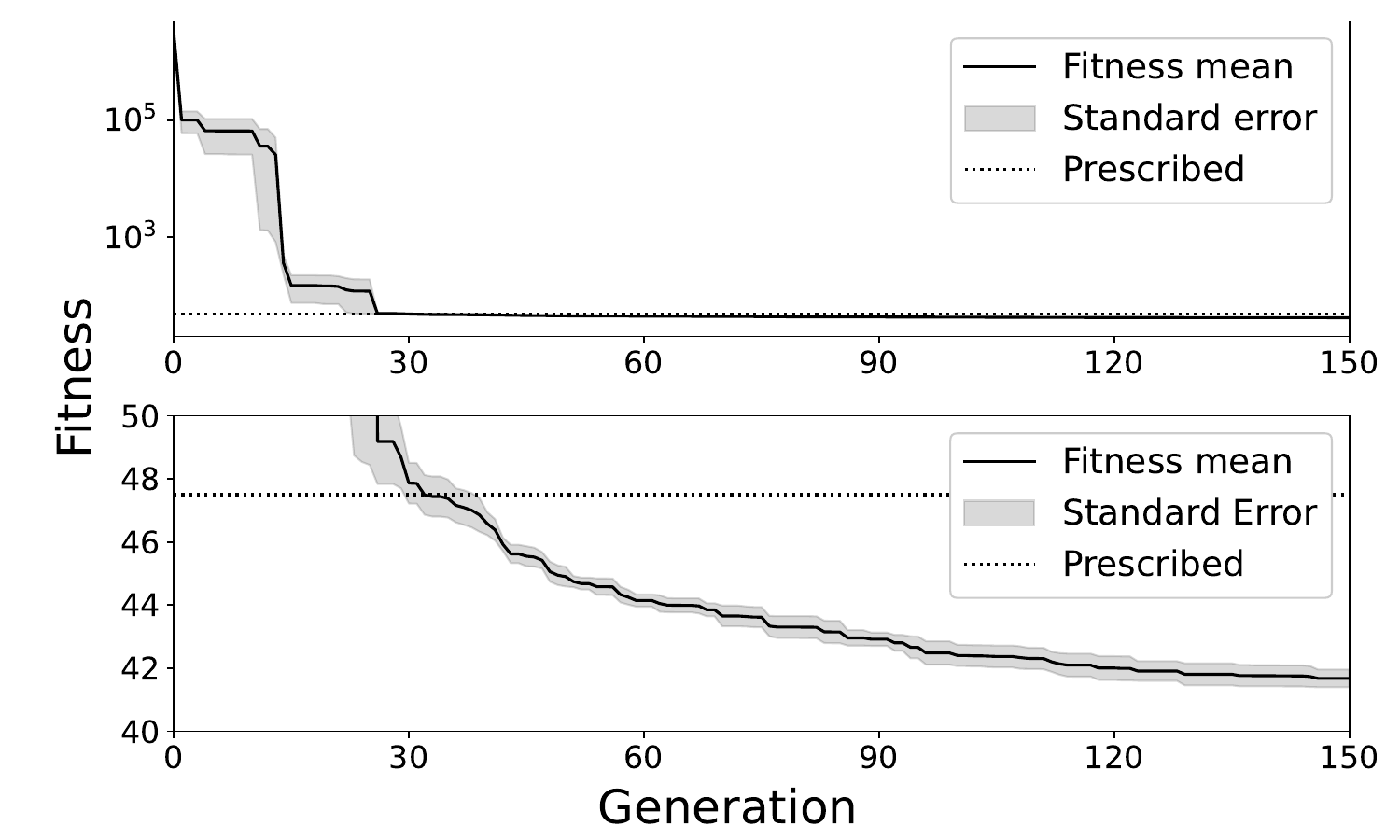} &
    \includegraphics[height=0.2\textheight,width=0.33\textwidth]{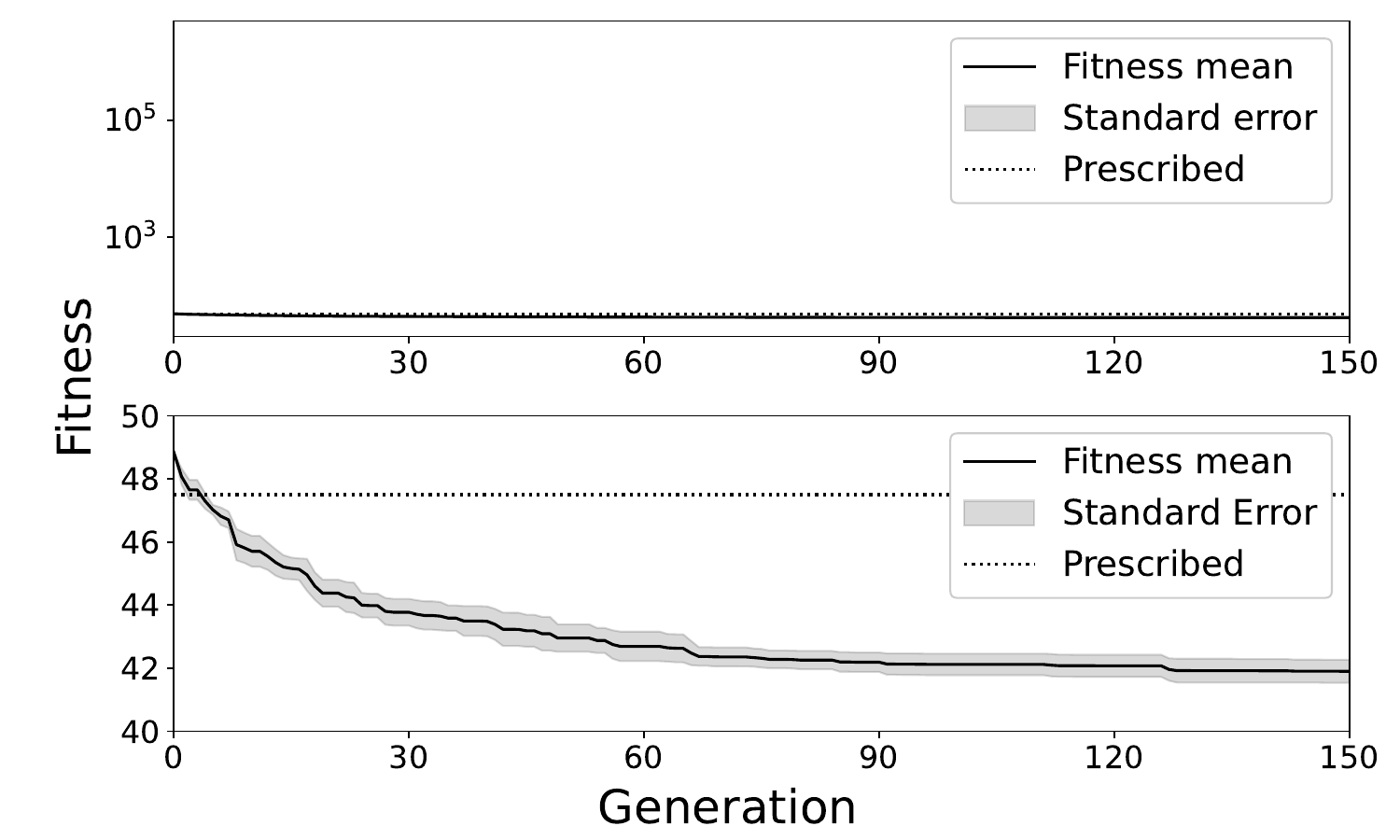} &
    \includegraphics[height=0.2\textheight,width=0.33\textwidth]{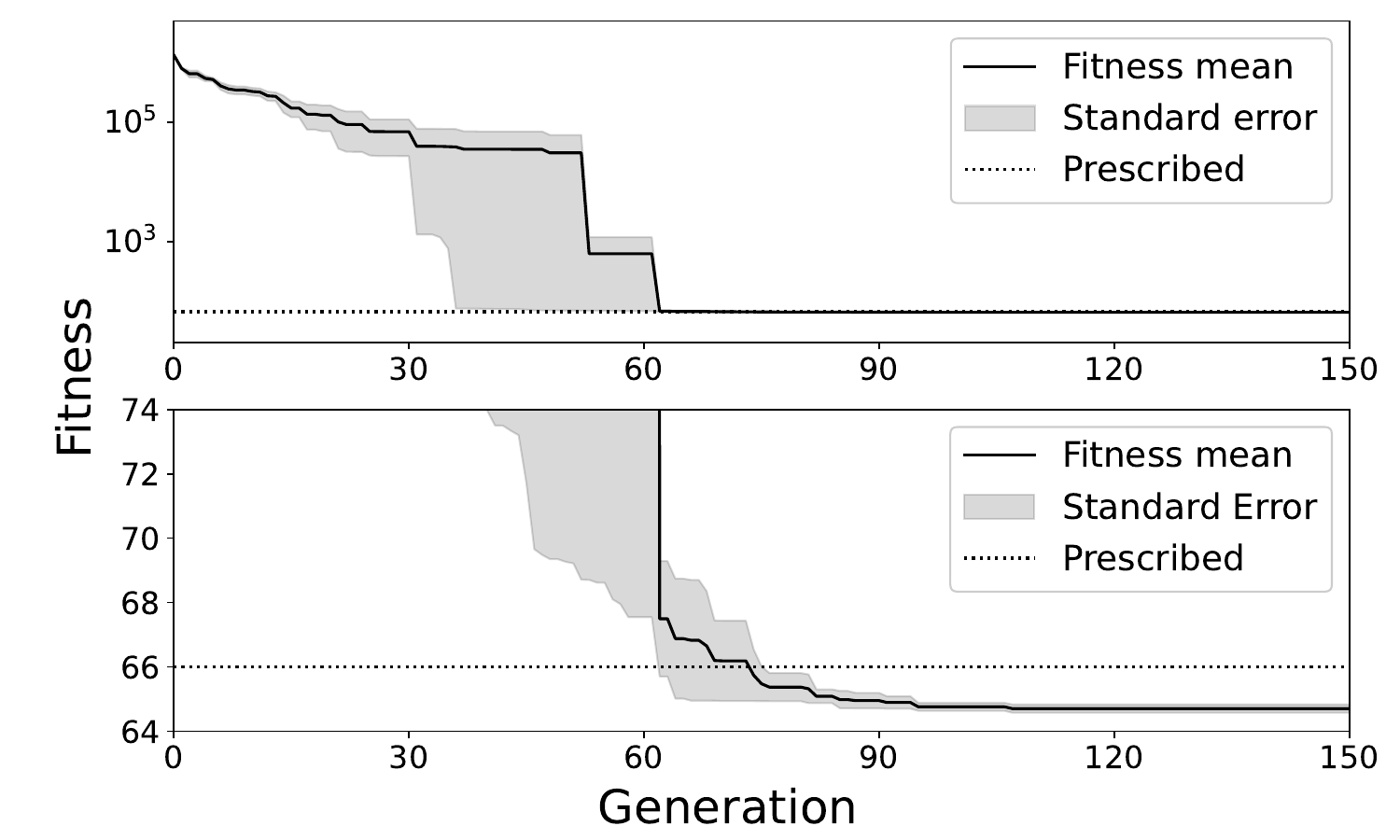}
    \\
    (a) $V_{1}$ & (b) $V_{2}$  & (c) $V_{3}$
  \end{tabular}
      }
  \caption{Transitions of the averaged objective function values of the best solutions for each generation.}
  \label{fig:fitness_transition}
\end{figure*}

Fig.~\ref{fig:fitness_transition} presents the transitions of the
averaged objective function value of the best individual per
generation.
The upper and lower graphs depict the same results but use different
vertical axis scales.
The upper graph tracks the overall trend of the objective function
values, while the lower graph focuses on those values after
constaraint violations have been resolved.
Because the post-resolution values exclude penalties, they represent
the maximum operating time of the arms (in seconds), and they appear
in the lower graph.
The solid line represents the average, and the shaded area denotes the
standard error.
On average, constraint violations were resolved by around generation
30, 5, 60 for $V_1$, $V_2$, and $V_3$, respectively.
Moreover, continued optimization yielded routes that finish
painting faster than prescribed time.
Adopting such routes can improve productivity by increasing daily
output and can allow for lower speeds of both arms and vehicles,
thereby enhancing paint quality.

The proposed method, which involved 150 generations of optimization to
create a detailed route, required an average processing time of about
240 minutes.

\subsection{Verification of practicality}
\subsubsection{Comparison with manually designed solution}
\label{ssec:comparison}

We compared the route on $V_3$ designed by human engineers with the
route generated by our method.
For the engineer-designed route, we calculated the objective function
value by forming a solution for the upper-layer subproblem (using the
same segment assignment and visiting schedule) and employing an
evaluation module that encompasses the lower-layer optimizer.

\begin{figure*}[t]
  \centering
      {\footnotesize
  \begin{tabular}{@{}c@{}c@{}}
    \vspace*{-8mm}
    \includegraphics[height=0.182\textheight,width=0.45\textwidth,trim={60mm 0mm 60mm 10mm},clip]{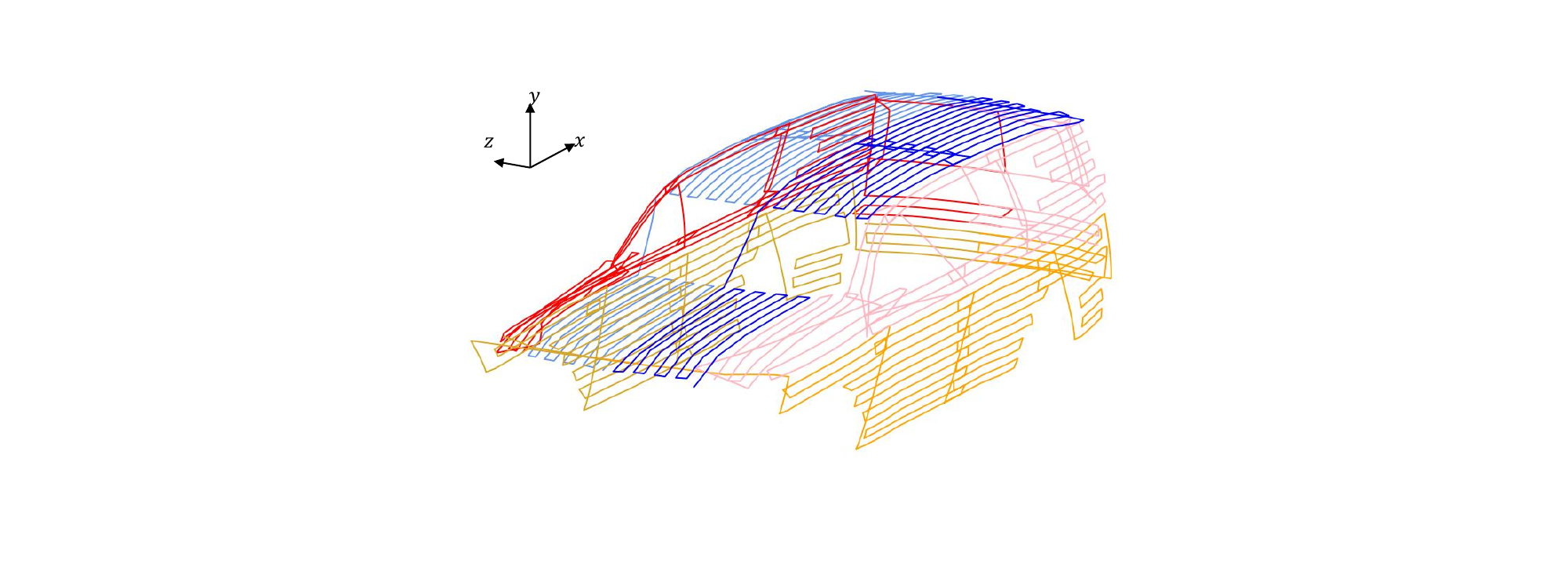} &
    \hspace*{3mm}
    \includegraphics[height=0.182\textheight,width=0.45\textwidth,trim={60mm 0mm 60mm 10mm},clip]{Fig/land_3Droute_03.pdf}\\
    (a) Engineer-designed route & (b) Route designed by the proposed method
  \end{tabular}
      }
  \caption{Detailed routes designed by the proposed method and engineers.}
  \label{fig:toyota_RouteCompare}
\end{figure*}

Fig.~\ref{fig:toyota_RouteCompare} illustrates the routes, and
Table~\ref{table:toyota_fitcompare} summarizes the comparison of the
objective function values and the number of violations for
Constraint~\ref{const:btm_to_up}.
Our method's route achieved a shorter working time than the engineer-designed
route while satisfying all strong constraints in
Table~\ref{table:constraints}, whereas the the engineer-designed route
incurred one violation of Constraint~\ref{const:btm_to_up}.

\begin{table}[tb]
  \caption{Comparison with the engineer-designed route for $V_3$.}
  \label{table:toyota_fitcompare}
    \centering
  \begin{tabular}{lrr}
    \hline
    & \multicolumn{1}{p{0.10\textwidth}}{Objective function Value $f$}
    & \multicolumn{1}{p{0.13\textwidth}}{Number of constraint violations}\\
    \hline
    Route by proposed method & 64.5 & 0\\
    Engineer-designed route  & 66.6 & 1 \\
    \hline
  \end{tabular}
\end{table}

\begin{figure}[tb]
  \centering
  \includegraphics[height=0.3\textwidth,width=0.49\textwidth]{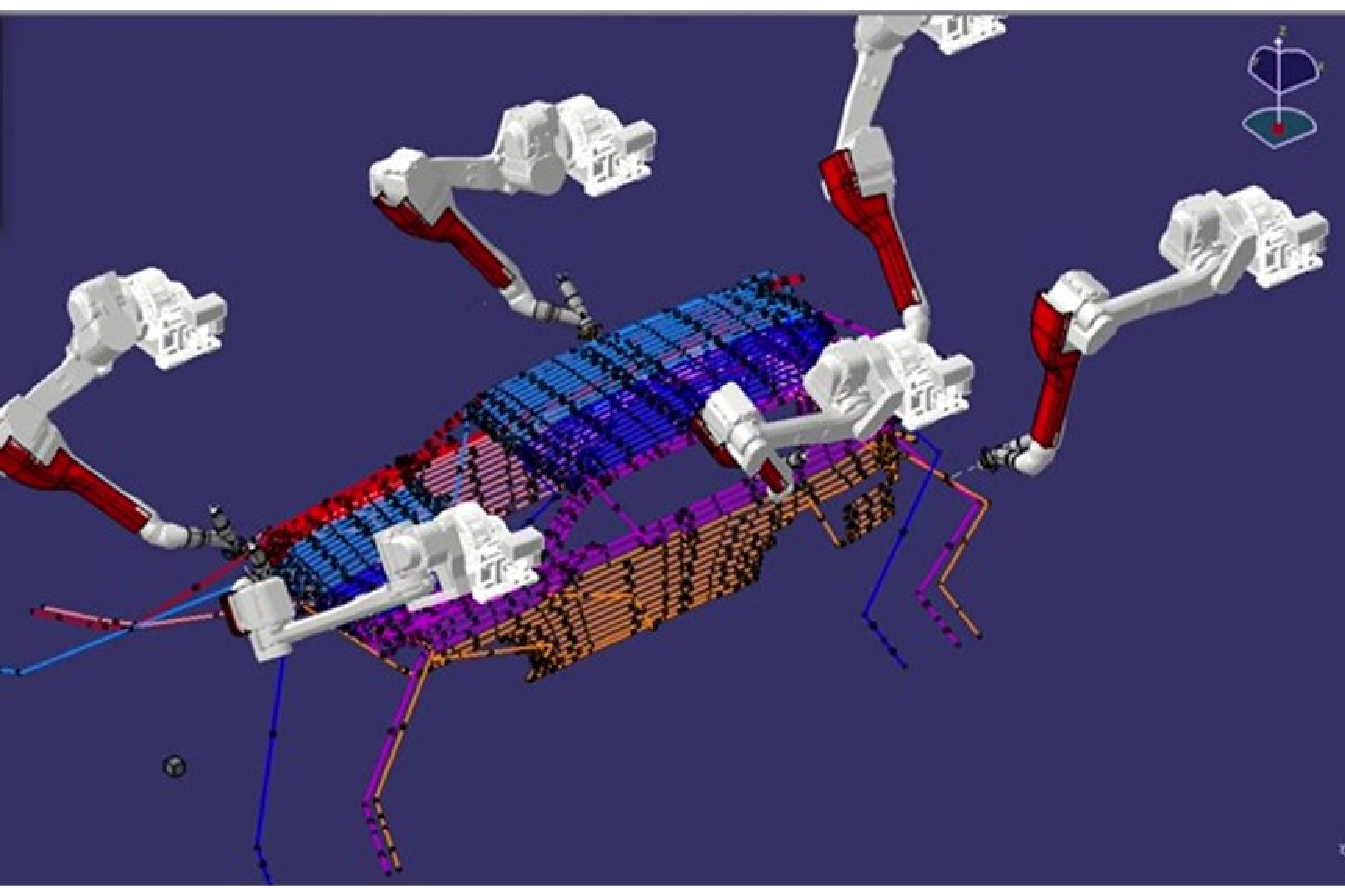}
  \caption{Result of applying the actual simulator used at $F_2$ to the route designed by the proposed method.}
  \label{fig:apply_toyota}
\end{figure}

\subsubsection{Application to actual production simulator}

We applied the route for $V_3$ designed using our proposed method to the
production simulator actually used in $F_2$.
Because our method computes only the arm head's trajectory, engineers
designed the intermediate joint motions based on the generated route
using the simulator.
Fig.~\ref{fig:apply_toyota} shows the results, which confirmed that the
route became feasible after modifying just one location.
This segment, located at the edge of the arm's range, arose because our
lower-layer optimizer simply approximates the arm's operating range as
spherical.

Furthermore, engineers estimated the time savings in painting route
design by introducing the proposed method at the production site.
Assuming the process requires seven weeks (one week for painting path
segment creation, five weeks for determining the arm's painting route,
and one week for interference checking), they found that our approach
could reduce the design time by roughly two weeks.
This improvement occurs because our method replaces the manual
trial-and-error steps formerly needed to meet the all constraints, and
it only requires minor adjustments when corrections are necessary.

\subsection{Ablation study}

Given the difficulty of a direct comparison with previous studies
presented in Section~\ref{sec:related_work}, we conducted an ablation
study to evaluate the performance effects of our method's
procedures---namely, initial solution generation and Repair
operators~\ref{repair:btm_to_up} and \ref{repair:less_arm}---%
tailored to vehicle painting path design.
5performance, we conducted an ablation study.
This experiment employed $V_3$, which includes a roof susceptible to
collisions, and compared eight different methods $M_1$ through $M_8$
configured as shown in Table~\ref{table:compare_method}.
Other experimental conditions followed Section~\ref{ssec:setup}.

\begin{table*}[t]
  \caption{Models compared in the ablation study.}
  \label{table:compare_method}
  \centering
  \begin{tabular}{l|cccccccc}
  \hline
    Method                                 
    & $\rm M_1$ & $\rm M_2$ & $\rm M_3$ & $\rm M_4$ & $\rm M_5$ & $\rm M_6$ & $\rm M_7$ & $\rm M_8$ \\  \hline
    Initial population generation          &  &\checkmark&  &  &\checkmark&\checkmark&  &\checkmark\\
    Repair operator~\ref{repair:btm_to_up} &  &  &\checkmark&  &\checkmark&  &\checkmark&\checkmark\\
    Repair operator~\ref{repair:less_arm}  &  &  &  &\checkmark&  &\checkmark&\checkmark&\checkmark\\
  \hline
  \end{tabular}

\end{table*}

\begin{figure*}[t]
  \centering
      {\small 
  \begin{tabular}{@{}c@{}c@{}}
    \includegraphics[height=0.24\textheight,width=0.48\textwidth]{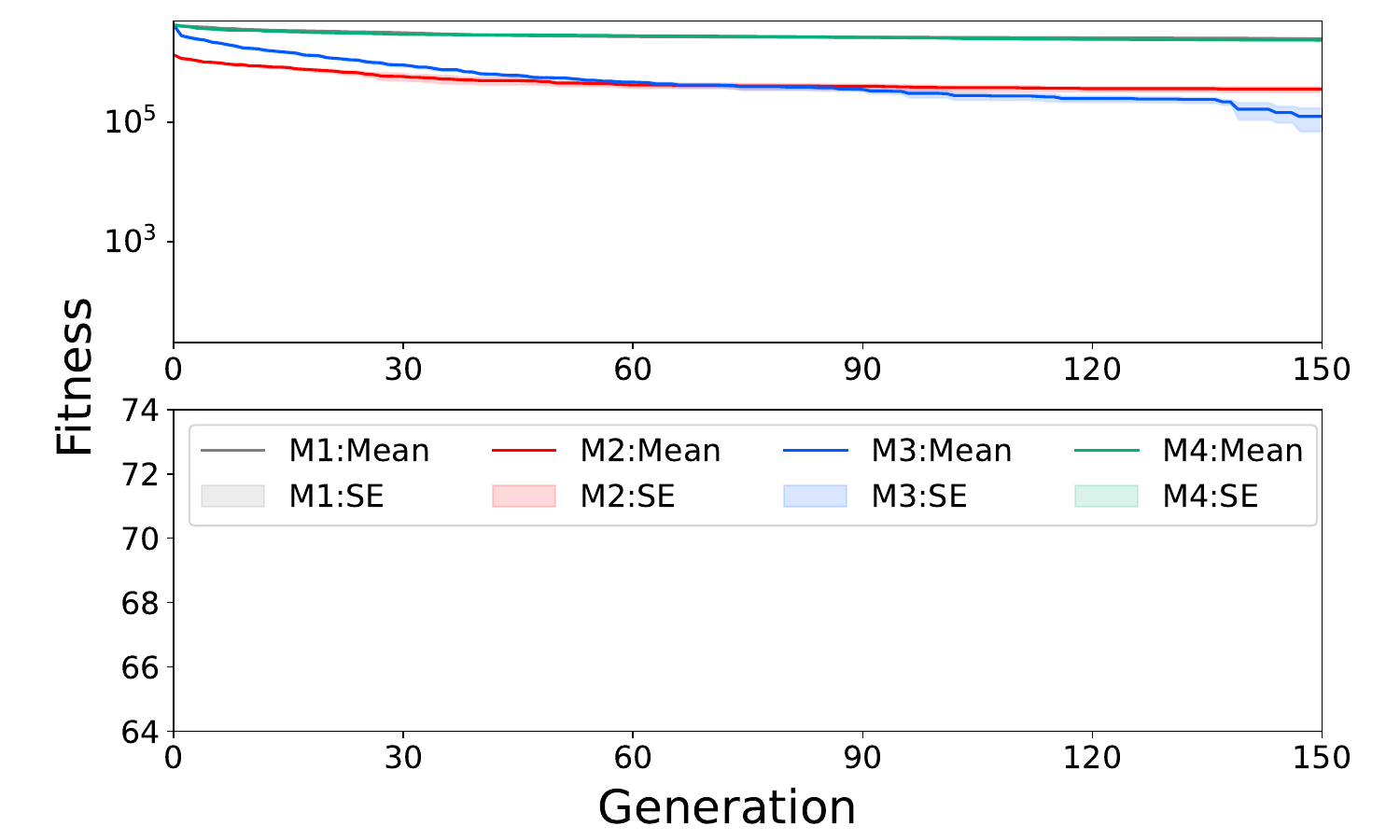} &
    \hspace*{3mm}
    \includegraphics[height=0.24\textheight,width=0.48\textwidth]{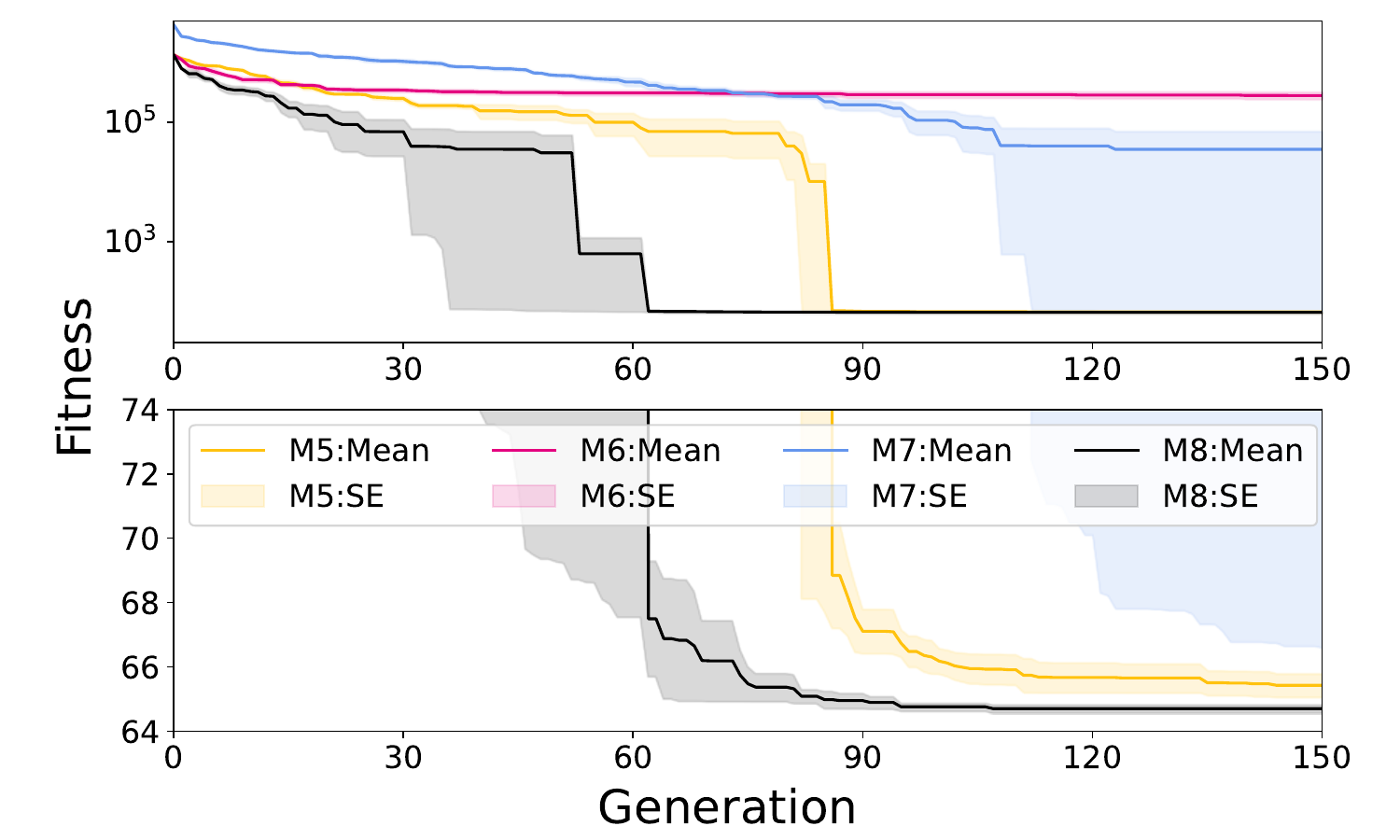}\\
    (a) $M_1$, $M_2$, $M_3$, and $M_4$ & (b) $M_5$, $M_6$, $M_7$, and $M_8$
  \end{tabular}
      }
      \caption{Comparison of transitions of the objective function values in the ablation study.}
  \label{fig:fitness_transition_ablation}
\end{figure*}

Fig.~\ref{fig:fitness_transition_ablation} displays the transitions of
objective function values for each method.
As in Fig.~\ref{fig:fitness_transition}, the upper graph represents 
the overall value changes, while the lower graph shows values after 
constraint violations have been resolved.
The solid lines indicate the averaged values, and the lighter shaded
areas denote the standard error.

$M_8$ (full method), which integrates improvements in both initial
solution generation and repair operators, produced the best objective
function value and satisfied all constraints in the fewest generations
compared to other methods.
Furthermore, when comparing methods $M_2$, $M_3$, and $M_4$---each
incorporating one improvement---it turns out that $M_3$, which only
includes Repair operator~\ref{repair:btm_to_up}, occasionally yielded
routes that met all constraints (except Constraint 5 regarding arm
operating time) and achieved the lowest evaluation value among
$M_2$, $M_3$, and $M_4$.
This suggests that, among the three enhancements, Repair
operator~\ref{repair:btm_to_up} exerted the most significant influence
on the objective function value.

\begin{figure*}[t]
  \centering
      {\small 
  \begin{tabular}{@{}c@{}c@{}}
    \includegraphics[height=0.14\textheight,width=0.42\textwidth,trim={40mm 10mm 40mm 5mm},clip]{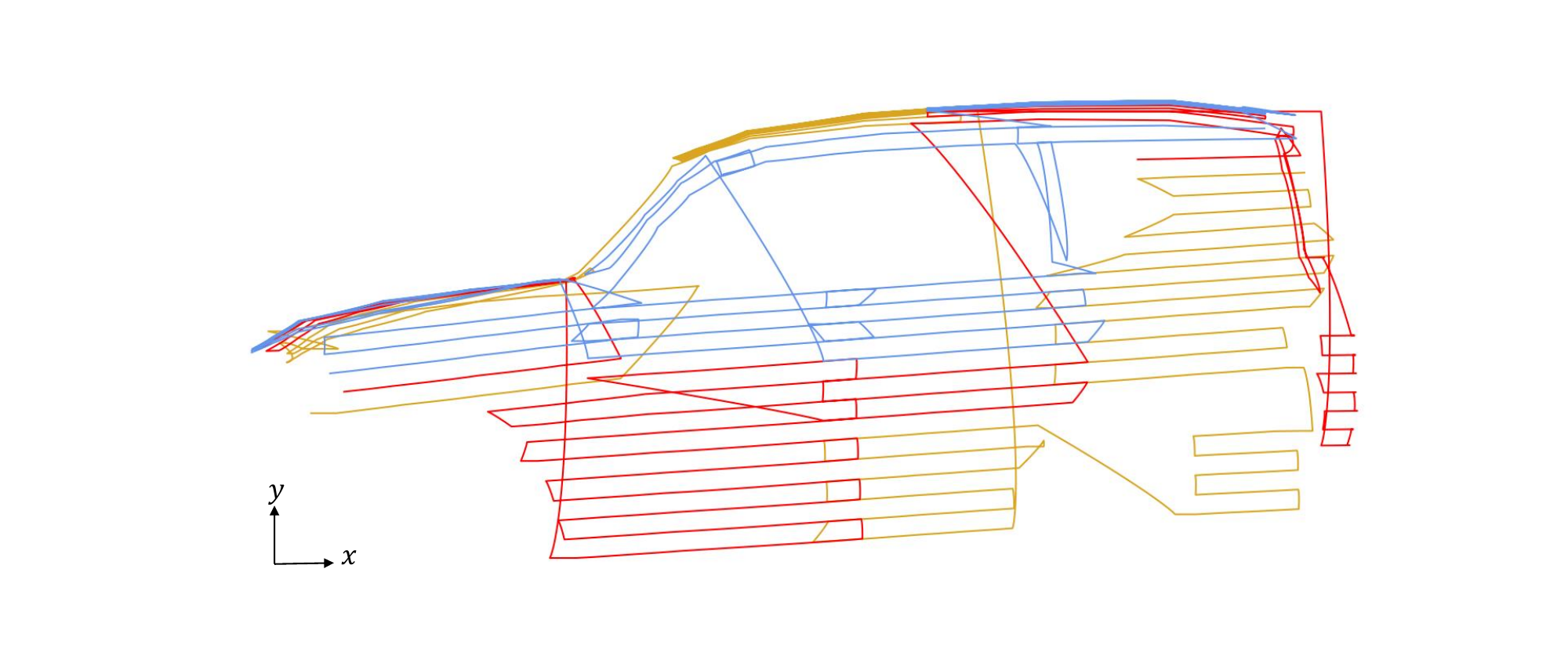} &
    \hspace*{15mm}
    \includegraphics[height=0.14\textheight,width=0.42\textwidth,trim={40mm 10mm 40mm 5mm},clip]{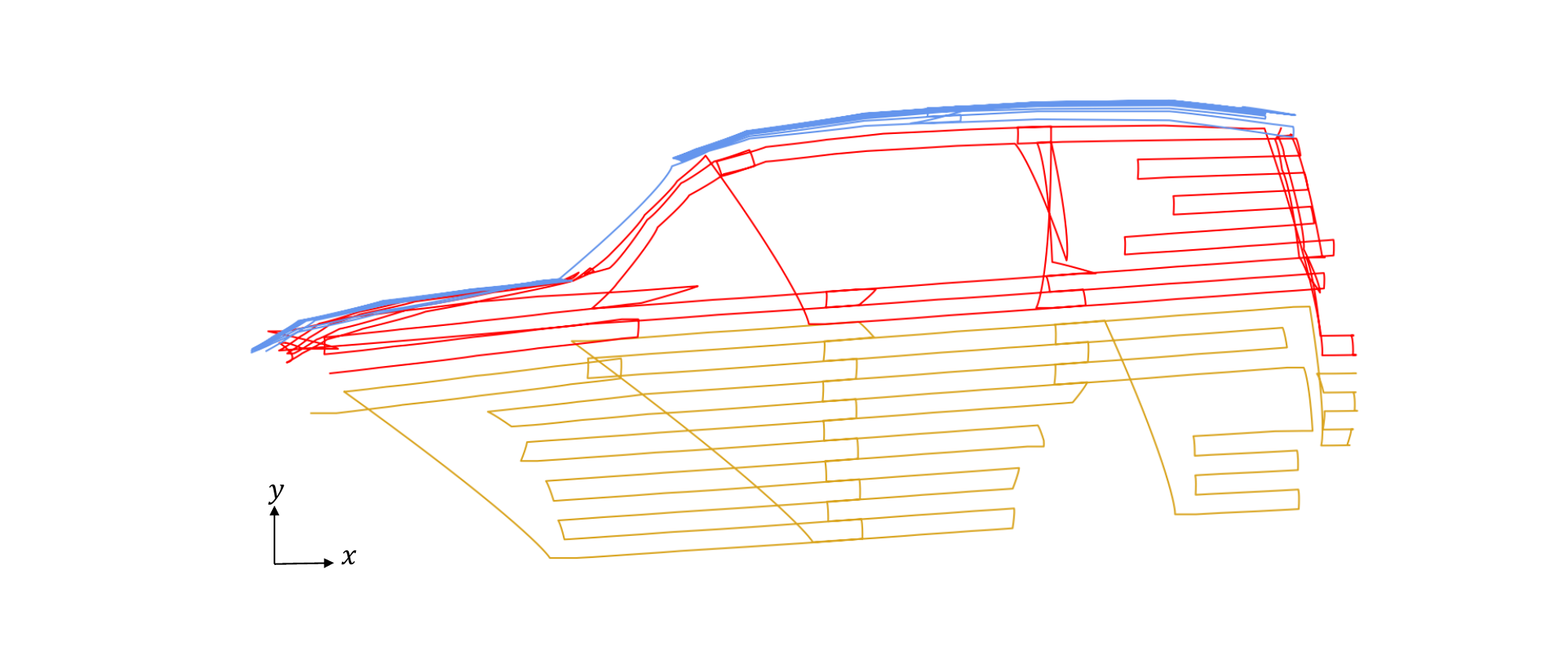}\\
    (a) Route on side and roof panels designed by $M_7$. & (b) Route on side and roof panels designed by $M_8$. 
  \end{tabular}
      }
      \caption{Comparison of detailed routes for side panels with and
        without the introduction of an initial population generation
        procedure shown in Section~\ref{ssec:init_pop}.}
  \label{fig:init_compare}
\end{figure*}

Among the eight methods, only $M_5$, $M_6$, and $M_8$ produced at
least one trial with routes that satisfied all
constraints. 
Specifically, $M_5$ succeeded in 4 out of 5 trials, whereas $M_6$ and
$M_8$ achieved feasibility in every trial. 
Moreover, comparing the consistently feasible methods revealed that
$M_8$ eliminated constraint violations about 20 generations earlier
and yielded lower evaluation values. 
These results indicate that while Repair
operator~\ref{repair:less_arm} and improvements in initial solution
generation alone did not guarantee fully compliant routes, combining
them with Repair operator~\ref{repair:btm_to_up} enhanced performance
more than applying repair operations alone.
In particular, incorporating Repair operator~\ref{repair:less_arm}
accelerated the resolution of constraint violations, and adopting a
horizontally aligned initial solution (as illustrated in
Fig.~\ref{fig:init_compare}) led to final routes that were both
horizontally uniform and robustly met all constraints.

\section{Conclusion}
\label{sec:conclusion}

This paper proposed a method for designing painting routes for multiple
robot arms on vehicle bodies.
We formulated the problem as a hierarchical optimization problem, 
with the upper-layer subproblem---responsible for assigning
painting regions to arms and determining the painting sequence---and
the lower-layer subproblem that designs the detailed route for each
arm.
This decomposition enables the use of metaheuristics, such as
evolutionary computation, to address both the robot arm constraints and
the unique constraints of the painting process.
Experiments with three vehicle types produced by two factories
confirmed that our method successfully designed routes that meet all
constraints.
Moreover, our method reduced the overall design process by
approximately two weeks.
In the future, we plan to incorporate multi-objective optimization to
design diverse routes suited to various production environments.

\bibliographystyle{unsrt}
\bibliography{access}

\begin{IEEEbiography}
[{\includegraphics[width=1in,height=1.25in,clip,keepaspectratio]{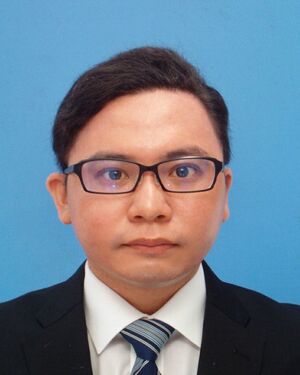}}]
{YUYA NAGAI} 
received his Bachelor's degree in Engineering from
Kagoshima University, Japan in 2023. 
He is currently a master course student of Department of Engineering,
Graduate School of Science and Engineering, Kagoshima University.
His research interest includes Optimization.
\end{IEEEbiography}

\begin{IEEEbiography}
[{\includegraphics[width=1in,height=1.25in,clip,keepaspectratio]{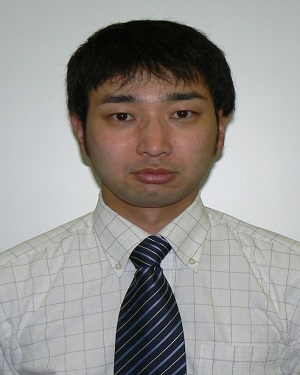}}]
{HIROMITSU NAKAMURA}
He received his Master of Engineering from Kagoshima University in
2000.  He worked at Mazda Motor Corporation from 2000 to 2006.
Subsequently, he joined TOYOTA AUTO BODY Research \& Development
CO.,LTD. and is currently a member of the Digital Engineering Div.  He
mainly works on IT and digital-related tasks related to production
technology for automotive welding and painting processes..  He is a
member of the Society of Automotive Engineers of Japan.
\end{IEEEbiography}

\begin{IEEEbiography}
[{\includegraphics[width=1in,height=1.25in,clip,keepaspectratio]{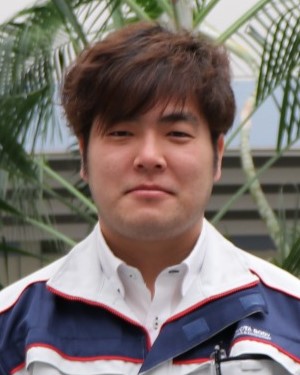}}]
{NARITO SHINMACHI}
He graduated from Miyakonojo Technical High School in 2009.
Subsequently, he joined TOYOTA AUTO BODY Research \& Development
CO.,LTD. and is currently a member of the Digital Engineering Div.  He
mainly works on IT and digital-related tasks related to production
technology for automotive welding and painting processes.  He is a
member of the Society of Automotive Engineers of Japan.
\end{IEEEbiography}

\begin{IEEEbiography}
[{\includegraphics[width=1in,height=1.25in,clip,keepaspectratio]{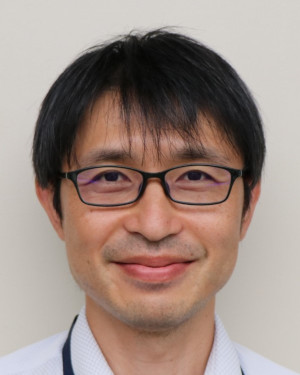}}]
{YUTA HIGASHIZONO}
He received his Ph. D. degree in Science from University of Tsukuba in
2008. He worked at Kyushu University as a Research Fellow of the
Japanese Society for the Promotion of Sciences (JSPS) from 2008 to
2010.  Subsequently, he joined TOYOTA AUTO BODY Research \& Development
CO.,LTD. and is currently a project manager of the Digital Engineering
Div.  He belongs to a department that promotes
industry-government-academia collaboration and promotes work to solve
automotive issues in collaboration with universities and research
institutes.  He is a member of the Society of Automotive Engineers of
Japan.
\end{IEEEbiography}

\begin{IEEEbiography}
[{\includegraphics[width=1in,height=1.25in,clip,keepaspectratio]{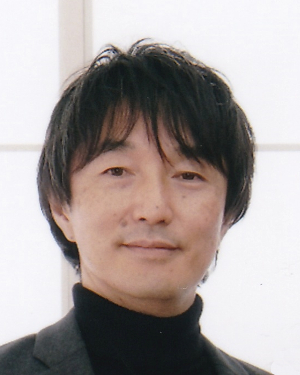}}]
{Satoshi Ono}
received his Ph.D. degree in Engineering at University of
Tsukuba in 2002. He worked as a Research Fellow of the Japanese
Society for the Promotion of Sciences (JSPS) from 2001 to
2003. Subsequently, he joined Department of Information and Computer
Science, Graduate School of Science and Engineering, Kagoshima
University as a Research Associate.  He is currently a Professor in
Department of Information Science and Biomedical Engineering in
Kagoshima University.  
He received JSAI Annual Conference Award 2023, IWAIT2020 best paper
award, TAAI2019 excellent paper award, IPSJ Yamashita SIG research
award 2012, etc. He is a member of IEEE, IPSJ, IEICE, and JSAI.  His
current research focuses on evolutionary computation, machine
learning, and their applications to real world problems.
\end{IEEEbiography}

\EOD

\end{document}